\documentclass[lettersize,journal]{IEEEtran}
\usepackage{amsmath,amsfonts}
\usepackage{tikz}
\usetikzlibrary{tikzmark,calc}

\usepackage{array}
\usepackage[caption=false,font=normalsize,labelfont=sf,textfont=sf]{subfig}
\usepackage{textcomp}
\usepackage{stfloats}
\usepackage{url}
\usepackage{verbatim}
\usepackage{graphicx}
\usepackage[draft]{hyperref}
\usepackage{url}
\usepackage{amssymb}
\usepackage{algorithm}
\usepackage{mathtools}

\usepackage{algpseudocode}
\usepackage{booktabs}
\usepackage{multirow}
\usepackage{multicol}

\usepackage{zref-user, zref-xr}
\usepackage{pifont}
\usepackage{colortbl}
\usepackage[numbers]{natbib}

\def\BibTeX{{\rm B\kern-.05em{\sc i\kern-.025em b}\kern-.08em
    T\kern-.1667em\lower.7ex\hbox{E}\kern-.125emX}}
\usepackage{balance}
\begin{document}
\title{Adversarial Video Promotion Against Text-to-Video Retrieval}

\author{Qiwei~Tian,~\IEEEmembership{Member,~IEEE,}
Chenhao~Lin\IEEEauthorrefmark{1},~\IEEEmembership{Member,~IEEE,}
Zhengyu~Zhao\IEEEauthorrefmark{1},~\IEEEmembership{Member,~IEEE,}
Shuai~Liu,~\IEEEmembership{Member,~IEEE,}
Qian~Li,~\IEEEmembership{Member,~IEEE,}
and~Chao~Shen,~\IEEEmembership{Fellow,~IEEE} \\

\IEEEauthorblockA{Xi'an Jiaotong University, Xi'an, China}

\thanks{\IEEEauthorrefmark{1}Corresponding authors.}
\thanks{Qiwei Tian, Chenhao Lin, Zhengyu Zhao, Qian Li and Chao Shen are with the Faculty of Electronic and Information Engineering, Xi'an Jiaotong University, Xi'an 710049, China (Email: michaeltqw@stu.xjtu.edu.cn; linchenhao@xjtu.edu.cn; zhengyu.zhao@xjtu.edu.cn; qianlix@xjtu.edu.cn; chaoshen@mail.xjtu.edu.cn).}
\thanks{Shuai Liu is with the School of Software Engineering, Xi'an Jiaotong University, Xi'an 710049, China (Email: sh\_liu@mail.xjtu.edu.cn) }
}

\markboth{Journal of \LaTeX\ Class Files,~Vol.~18, No.~9, September~2020}%
{How to Use the IEEEtran \LaTeX \ Templates}

\maketitle

 \begin{abstract}
Thanks to the development of cross-modal models, text-to-video retrieval (T2VR) is advancing rapidly, but its robustness remains largely unexamined. Existing attacks against T2VR are designed to push videos \textit{away} from queries, i.e., suppressing the ranks of videos, while the attacks that pull videos \textit{towards} selected queries, i.e., promoting the ranks of videos, remain largely unexplored. These attacks can be more impactful as attackers may gain more views/clicks for financial benefits and widespread (mis)information. To this end, we pioneer the first attack against T2VR to promote videos adversarially, dubbed the Video Promotion attack (ViPro). We further propose Modal Refinement (MoRe) to capture the finer-grained, intricate interaction between visual and textual modalities and enhance black-box transferability. Comprehensive experiments cover \textit{2} existing baselines, \textit{3} leading T2VR models, \textit{3} prevailing datasets with over 10k videos, evaluated under \textit{3} scenarios. All experiments are conducted in a multi-target setting to reflect realistic scenarios where attackers seek to promote the video regarding multiple queries simultaneously. We also evaluated our attacks for defenses and imperceptibility. Overall, ViPro surpasses other baselines by over $30/10/4\%$ for white/grey/black-box settings on average. Our work highlights an overlooked vulnerability, provides a qualitative analysis on the upper/lower bound of our attacks, and offers insights into potential counterplays. Code is available at \hyperlink{https://github.com/michaeltian108/ViPro}{https://github.com/michaeltian108/ViPro}.
\end{abstract}

\section{Introduction}
\label{sec:intro}
\IEEEPARstart{T}ext-to-video retrieval (T2VR) has thrived in recent years, driven by advances in vision–language models (VLM) \cite{ref:clip2v,ref:Clip,ref:Clip-vip,ref:blip}, making it a vital tool for users to search, discover, and consume video content across. While many T2VR models \cite{ref:HBI,ref:singularity,DRL,C4V} have achieved strong performance on prevailing datasets such as MSR-VTT \cite{ref:MSRVTT}, DiDeMo \cite{ref:DDM}, and ActivityNet \cite{ref:ANT}, their robustness to adversarial attacks remains largely unexplored. For example, as the only existing attack against T2VR, \citet{VTRA} proposed an adversarial attack against T2VR models to \textbf{suppress video ranks} by pushing videos away from queries and vice versa under black-box and white-box settings. However, attacks that \textbf{promote videos to top positions} can be more hazardous, as they allow attackers to gain more views/clicks for financial benefits and widespread (mis)information. Additionally, such manipulation can trigger a snowball effect: once the manipulated videos gain initial popularity through adversarial promotion, the platform's recommendation/retrieval algorithm may further amplify their exposure. Recognizing this overlooked threat, we investigate this vulnerability to video promotion attacks as a critical step towards building robust and trustworthy T2VR systems.

\begin{figure}[!t]
\centering
  \includegraphics[width=0.9\linewidth]{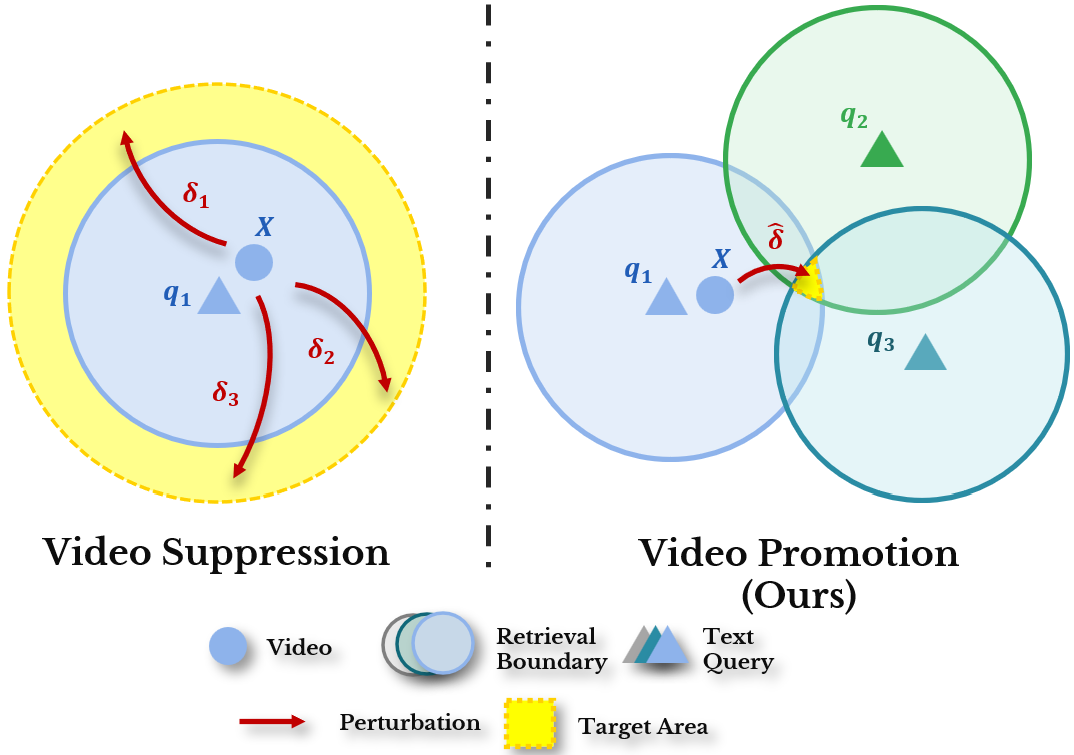}
\caption{
A simplified illustration for video suppression (left) and video promotion (right). The circle represents the retrieval boundary in which videos can be retrieved by the query (text). The yellow shape indicates the targeted area in which the perturbed videos must reach for successful attacks. \textbf{Video promotion is more challenging, as it seeks an intersection of the retrieval boundaries for all target queries, while video suppression only pushes videos outside the retrieval boundary.}
}
\label{fig:vl}
\end{figure}

To this end, we propose the first attack against T2VR to promote videos adversarially for the selected queries, dubbed the Video Promotion attack (ViPro). Furthermore, to reflect realistic user scenarios, we restrict the attack to candidate videos only, without modifying the text queries, as attackers cannot interfere with queries from other users. In Figure.~\ref{fig:vl}, we provide a hypothesized illustration to highlight the difference between existing attacks (video suppression) and ours (video promotion) intuitively. In the left figure, video suppression attacks succeed if the perturbation $\delta$ pushes the video $\mathbf{X}$ out of the retrieval boundary of $q_1$ (depicted as the blue circle), which is determined by the retrieval list length and the videos scattered around the query. Thus, the desired perturbation can be obtained by simply maximizing the distance between the video $\mathbf{X}$ and the query $q_1$, as illustrated by the red arrows ($\delta_1,\delta_2,\delta_3$). In contrast, video promotion attacks (right figure) aim to push $\mathbf{X}$ into the small overlapping area (yellow shape) of the retrieval boundaries for all target queries (e.g., $q_1,q_2,q_3$). Thus, finding the optimal $\hat\delta$ requires pre-knowledge of model embeddings and the distribution of target queries. 

To overcome the above challenges for better transferability, we further propose \textbf{Mo}dality \textbf{Re}finement (MoRe) to guide perturbation towards target areas through finer-grained optimization. MoRe first conducts \textit{Temporal Clipping} to group temporally similar frames into clips, and then applies \textit{Semantic Weighting} to resolve potential gradient conflicts for intra-modality and inter-modality through frame-to-frame and frame-to-query similarities, respectively. Consequently, MoRe provides focused guidance to push the video towards the targeted queries for better transferability. We validate the effectiveness of MoRe through ablation studies in Section.\ref{sec:as}. 

Overall, we conduct experiments on \textbf{3} open-sourced leading models for MSR-VTT-1k~\cite{ref:MSRVTT}, DRL~\cite{DRL}, Cap4Video~\cite{C4V}, and Singularity~\cite{ref:singularity} (ranked \#11/\#17/\#30\footnote{Using the leaderboard from \href{https://paperswithcode.com/sota/video-retrieval-on-msr-vtt-1ka}{PapersWithCode.com}.}). We define the white-box, grey-box, and black-box scenarios for ViPro and evaluate its effectiveness and transferability across all three models. For dataset-wise evaluations, we test on \textbf{three} prevailing datasets using the Singularity model—ActivityNet~\cite{ref:ANT}, DiDeMo~\cite{ref:DDM}, and MSR-VTT. As for baselines, we adopt Co-Attack\cite{Coattack} and SGA\cite{SGA} from the image-to-text domain for comparison, due to the lack of reproducible T2VR-specific attacks. Lastly, we further assess robustness under defences using JPEG compression~\cite{jpeg} (image-based) and Temporal Shuffling~\cite{ts} (video-based). A user study with 17 experts on 43 groups of videos is also conducted for human imperceptibility. \textbf{Across all baselines across all datasets, models, and settings, ViPro consistently outperforms other baselines.}  Finally, we validate our theoretical motivation through thorough experiments and provide detailed analysis in Section~\ref{sec:discussion}.

In sum, ViPro exhibits superior effectiveness, generalizability, and transferability over existing baselines, achieving an averaged $30/10/4\%$ lead for white/grey/black-box settings. Our main contributions are listed as follows:

\begin{itemize}
    \item We introduce a realistic and impactful attacking paradigm for text-to-video retrieval (T2VR) models where attackers promote video ranks for selected queries adversarially. We thus pioneer the Video Promotion Attack (ViPro) as the first attack targeting such vulnerability.
    
    \item We propose Modality Refinement (MoRe) to group temporally similar frames into clips and apply semantical weighting using frame-to-frame and frame-to-query similarities for enhanced black-box transferability. 
    
    \item We conduct thorough experiments and validate the superiority of ViPro over baselines and provide a qualitative analysis of the upper and lower bounds for all attacks. 
\end{itemize}
\section{Related Work}
\label{sec:RW}

\noindent\textbf{Vision-Language Models.}
Vision-Language models (VLMs), as suggested by the name, require training in both textual and visual encoders. Textual encoders have witnessed the advancement in natural language processing, which progressed from Word2Vec\cite{ref:nlp1} to Bert-base \cite{ref:Bert, ref:nlp2} and GPT-based\cite{ref:nlp3} encoders. Besides, visual encoders\cite{ref:ViT,ref:vit1,ref:vit2,ref:vit3,ref:vit4,ref:vit6} also advance significantly, surpassing CNN on numerous tasks in the image domain. In light of that, pre-trained VLMs\cite{ALBEF, ref:Clip,ref:Beit,ref:vidit1,ref:vidit2,ref:vidit3} have achieved promising performance on various multi-modal tasks with proper fine-tuning. 

\noindent \textbf{Text-to-Video Retrieval.}
Following the paradigm of VLMs, text-to-video retrieval (T2VR) models also use various off-the-shelf encoders to extract features from videos and their captions\cite{ref:t2vr1,ref:t2v2, ref:t2v4,ref:t2v5,ref:t2v6}. In recent works, CLIP \citet{ref:Clip} proposes contrastive learning to align visual representations with textual features to achieve an impressive cross-modal retrieval performance. Following this work, many T2VR models have adopted contrastive learning. 
For example, CLIP-ViP \cite{ref:Clip-vip} uses pre-trained image-text models and proposes an Omnisource Cross-modal Learning method to align videos to captions and subtitles in a contrastive learning way. \textcolor{black}{More recently, \citet{tian2024towards} proposed EERCF, which uses a multi-granularity visual feature learning with a two-stage retrieval structure, achieving similar performance with much less computational cost. As for cross-modal interaction, EERCF adopts a parameter-free text-gated interaction block (TIB) to handle text-video alignment. Later, \citet{TeachClip} proposed TeachCLIP which uses the text-video similarity from a frozen teacher network as the soft labels to train a student network, achieving promising results compared to several CLIP-based models. TeachCLIP also uses a modular block called Attentional frame-Feature Aggregation (AFA) to allocate weights for coarse-/fine-grained frame features for efficient knowledge distillation. } In sum, the architecture of all existing T2VR models can be modelled similarly as ``two-plus-one'': two encoders to encode visual and textual inputs, plus one module responsible for cross-modal interaction. All prevailing models can be modelled into such a paradigm, such as Singularity\cite{ref:singularity} DRL\cite{DRL}, and Cap4Video\cite{C4V}. Despite the flexibility of such paradigms, the adoption of pre-trained encoders induces risks of adversarial attacks from knowledgeable attackers. Our work shows that ViPro remains effective with the absence of cross-modal modules, even for completely black-box scenarios.

\noindent \textbf{Adversarial Attacks on Unimodal Retrieval.}
Despite numerous explorations on uni-modal and multi-modal retrieval \cite{QAIR,VTRA,Coattack,SGA, Dair}, existing works largely focus on the untargeted paradigm, i.e., suppressing the ranks of videos regarding their corresponding queries. For example, \citet{QAIR} proposed a novel query-based adversarial attack to completely inverse ranks for image retrieval. \citet{Dair} proposed a query-efficient adversarial attack that efficiently suppresses the ranks of manipulated images. Among them, very limited literature has focused on targeted attacks against retrieval models. Specifically, \citet{OA} proposed an Order Attack against image retrieval that manipulates the order of the retrieved results by manipulating the query images. 

\noindent \textbf{Adversarial Attacks on Multimodal Retrieval.}
Research on adversarial attacks on multimodal retrieval has primarily focused on the image domain\cite{ref:AA_on_image_caption,ref:AAonVLP, ref:evAdvRob, ref:multi-modalAdv, ref:GB_AA_on_I2T}. Recent studies \cite{Coattack,SGA,VLA,SA} have demonstrated transferable cross-modal adversarial attacks against pre-trained VLM on tasks including text-to-image retrieval, VQA, etc. \textcolor{black}{These works have shown the impressive effectiveness of using tailored augmentation to boost transferability. Similarly, \citet{DRA} and its extension \cite{DRA_ext} utilized triangular relationship to further push images away from the paired texts. \citet{MAA} proposed to use a frequency enhancement to boost cross-modal transferability. \citet{GLEAM} further refined such augmentation through global-local guidance called GLEAM, surpassing existing SOTA.}

For attacks against T2VR, \citet{VTRA} has proposed an adversarial attack against models such as Clip4Clip\cite{c4c}, Frozen\cite{FT}, and BridgeFormer\cite{BF}, using the MSR-VTT\cite{ref:MSRVTT} and DiDeMo\cite{ref:DDM} datasets. However, this work mainly focused on the untargeted setting to suppress videos ranks w.r.t. their corresponding queries. 
Another similar work was conducted by \citet{trojan}. Specifically, the author proposed a Trojan-horse attack (THA) against text-to-image retrieval that was the first rank promotion adversarial targeted on multiple modalities. Specifically, THA added adversarial QR patches into images to promote the rank of the images w.r.t. target text queries. However, this work only focused on text-to-image retrieval and used an adversarial QR patch as perturbations, which are too obvious and not applicable to videos. Lastly, THA is only targeted at \textbf{1} query per image, which is much less challenging than our ViPro, which targets multiple queries simultaneously. Consequently, we raise the importance of investigating targeted attacks against T2VR and propose ViPro as a realistic and impactful attacking paradigm.

\section{Methodology}
\label{sec:method}
\newcommand{\HighlightBlockf}[2]{
  \tikz[remember picture,overlay]{
    \draw[thick,black,rounded corners]
      ($(pic cs:#1)+(-0.5\linewidth,-0.4ex)$) rectangle
      ($(pic cs:#2)+(0.04\linewidth, -0.8ex)$);
  }
}

\newcommand{\HighlightBlocks}[2]{
  \tikz[remember picture,overlay]{
    \draw[thick,black,rounded corners]
      ($(pic cs:#1)+(-0.95\linewidth,-0.6ex)$) rectangle
      ($(pic cs:#2)+(0.85\linewidth, 1.8ex)$);
  }
}

\newcommand{\HighlightBlockt}[2]{
  \tikz[remember picture,overlay]{
    \draw[thick,black,rounded corners]
      ($(pic cs:#1)+(-0.68\linewidth, 4.8ex)$) rectangle
      ($(pic cs:#2)+(0.25\linewidth, -1.5ex)$);
  }
}

\subsection{Threat Model}
\label{sec:TM}
As illustrated in Table.\ref{tabl_as}, we first define attacking scenarios by categorizing victim models as white-box, grey-box, and black-box. Specifically, the grey-box setting accounts for the situation when the cross-modal interaction module remains unknown to the attackers.

For attackers, we hypothesize that attackers prudently target only semantically relevant queries to avoid obvious misalignment with unrelated queries, e.g., a food video occurs in a sports-related query. Similar to image classification, ViPro can be regarded as a targeted attack focusing on multiple semantically related queries, as we depicted in Figure.\ref{fig:vl}.

For queries, we follow existing works\cite {DRL,ref:singularity,c4c,C4V} and use the paired caption or category of the video as its query. For white-box and grey-box scenarios, attackers can access a subset of queries through the subtitles/descriptions of videos. (See details in Sec.\ref{sec:ES} and the Supplementary.) Specifically, for black-box attacks, attackers can only access the categories (domains) of videos and evaluate the attack accordingly. 

Lastly, we formalize the victim model as an open T2VR platform, allowing users to query or upload their videos (e.g., YouTube). Attackers can only launch attacks by uploading manipulated videos to acquire more views/clicks for potential financial gains and widespread (mis)information.

\begin{table}[t!]
     \caption{\label{tabl_as} Settings for attackers' per-knowledge in different scenarios.}
    \centering
   \resizebox{\linewidth}{!}{
   \begin{tabular}{c|c|c|c}
    \toprule
    \multicolumn{3}{c|}{\textbf{Attackers' Pre-knowledge}} & Attacking\\
    \cmidrule(lr){1-3}
    Caption/Category & Unimodel Encoders&  Cross-modal Interaction &   Scenarios \\
    \cmidrule(lr){1-4}
   Caption & \checkmark & \checkmark & White-box \\
    Caption & \checkmark & \ding{55} & Grey-box \\
    Category &  \ding{55} & \ding{55} & Black-box \\
    \bottomrule

    \end{tabular}}
\end{table}

\subsection{\textcolor{black}{Preliminaries}}

\noindent\textbf{\textcolor{black}{Task Definition}}. For a T2VR model, we define $\mathcal{V}$ and $\mathcal{T}$ as the visual and text encoders, respectively, with $\mathcal{H}$ being the cross-modality interaction, which could be a dot product, an MLP layer, or a transformer, etc. \textcolor{black}{$\mathcal{D}$ is the video corpus (candidates)}.
Formally, for an input query ${q}\in\mathbf{Q}$, the T2VR task is to rank all videos by their similarities w.r.t. $q$ and output a video list $\mathbb{X}_{q,L}$ that contains the top $L$ relevant video candidates $\mathbf{X}_i$ for the query:
\begin{equation}
\label{eq:retrie}
 \mathcal{H}\Big(\mathcal{T}({q}),\mathcal{V}(\mathcal{D})\Big) \rightarrow \mathbf{X}_{q,L} \coloneqq [\mathbf{X}_1,\mathbf{X}_{2},...,\mathbf{X}_{L}]_q 
\end{equation}
The performance of a model can be evaluated based on the number of corresponding videos within the retrieval list ${\mathbf{X}}_{q,L}$. 

\noindent\textbf{Retrieval Boundary}. We now formally define the mathematical definition of retrieval boundaries. Based on our hypothesis in Fig.\ref{fig:vl}, for a perturbed video to appear in the retrieval list ${\mathbf{X}}_{q,L}$, the video must cross the retrieval boundary of the chosen query. Such a boundary is determined by the length of the list $L$ and the spatial distribution of videos in the proximity. We use $L=1$ to consider top-1 videos only for simplification. Denoting the retrieval boundary of a query $q$ as a hyper-sphere $\Phi_q$ of radius $r$, $r$ is thus the distance towards its closest video around, denoted $\mathbf{X}_q$. Once the manipulated video enters $\Phi_q$, the video becomes closer than $\mathbf{X}_q$, replacing it as the new top-1 video. Thus, the radius of $\Phi_q$, $r$, is inversely proportional to the similarity of the closest video nearby. In other words, larger top-1 similarity means smaller $r$, making it harder for attacks to cross $\Phi_q$, or vice versa.

\noindent\textbf{Attack Objectives.}
Our goal is to push the candidate video $\mathbf{X}$ such that $\mathbf{X}$ appears at the top-1 position for all selected queries: 
\begin{equation}
    \label{eq:retrievability}
    {\mathbf{X}'} \in \mathbf{X}_{q,L}
    , \forall q \in \mathbf{Q}
\end{equation}
where ${\mathbf{X}}'\coloneqq\mathbf{X} + \delta$, with $\delta$ being the adversarial perturbations. Similar to other adversarial attacks, ViPro also requires visual constraints to keep the attack imperceptible to human eyes, i.e., $||\delta||_{p} \leq \epsilon$, where $\epsilon$ is the threshold for the $l_p$ norm. Note that only \textit{visual} modality is attacked.

\subsection{\textcolor{black}{Video Promotion Attack (ViPro)}}
\textcolor{black}{Overall, Video Promotion Attack (ViPro) consists of two modules: \textbf{Loss Function} and \textbf{Modality Refinement}. The former serves as the core optimization objective for all scenarios, while the latter is introduced for enhanced transferability in black-box scenarios. }

\subsubsection{Loss Function}
Due to the discreteness of $\mathbf{X}_{q,L}$, Eq.\ref{eq:retrievability} is not directly differentiable. We thus apply our retrieval boundary hypothesis above, which translates our objective into pushing $\mathbf{X}$ towards the overlap of the retrieval boundaries for all queries in $\mathbf{Q}$, i.e., $\mathbf{X}' \in \Phi_{q_1}\cap\Phi_{q_2}...\cap\Phi_{q_N}$. The optimal objective to achieve this is to increase the similarity between the candidate video $\mathbf{X}$ to exceed the similarity for all top-1 videos $\mathbf{X}_q \in\mathbf{X}_\mathbf{Q}$. However, this optimal objective requires access to $\mathbf{X}_\mathbf{Q}$, which is unrealistic. An approximation is to increase the similarity for all queries. Formally, denoting the vision/text feature as $\mathbf{F}_{\mathbf{X}}$ and $\mathbf{F}_{q}$, we define their cross-modality similarity $\mathbf{S}$ is defined below:

\begin{equation}
\label{eq:sim}
    \mathbf{S}=\sum^{\mathbf{Q}}_{q}Sim\big(F_{\mathbf{q}},F_\mathbf{X}\big)
\end{equation}
where $Sim(\cdot)$ refers to the cross-modal interaction to calculate the similarities between videos and queries. For white-box, $Sim(\cdot)$ will be the cross-modal module of victim models, while for grey-box attacks, it will be the cosine similarity. $F_\mathbf{X},F_{\mathbf{q}}$ refer to video and queries features.

Intuitively, a naive solution is to use the negative similarity scores as the loss function directly:
\begin{equation}
\label{eq:lnaive}
 \mathcal{L}_{neg}(\mathbf{S}) = - \mathbf{S}
\end{equation}
However, $\mathcal{L}_{neg}$ will lead to suboptimal results, especially for multi-target optimization, because it provides identical gradients (i.e., 1) for all targets. Thus, we propose to use exponential loss for optimal results:
\begin{equation}
\label{eq:lexp}
 \mathcal{L}_{exp}(\mathbf{S}) = \exp(-\mathbf{S})
\end{equation}
$\mathcal{L}_{exp}$ could provide adaptive gradients for different targets, i.e., larger gradients for lower $\mathbf{S}$ (farther targets) and lower gradients for larger $\mathbf{S}$ (nearer targets). Experimental results also validate the effectiveness of $\mathcal{L}_{exp}$.

\begin{figure*}[!t]
\centering
  \includegraphics[width=\linewidth]{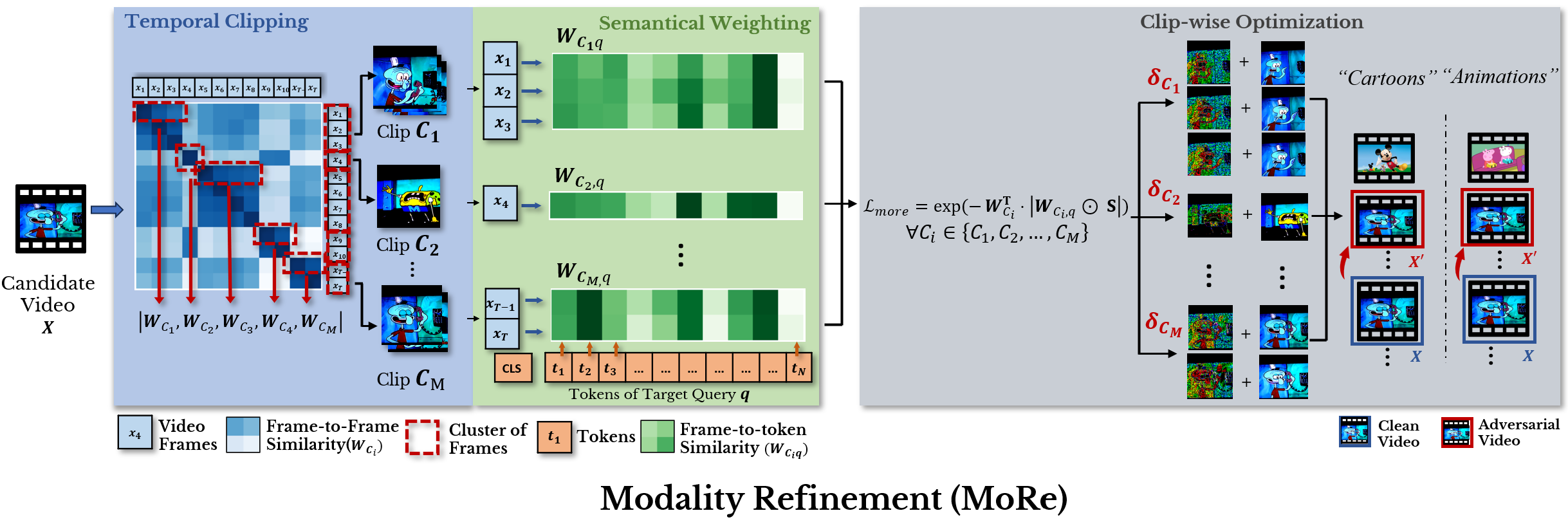}

\caption{\label{fig:pipline}
An illustration of Modality Refinement, i.e., Temporal Clipping and Semantical Weighting. (1) \textbf{Temporal Clipping:} Video frames are clustered into video clips $\mathbb{C}=[\mathbf{C}_1, ...,\mathbf{C}_M]$  based on frame-to-frame similarity $\mathbf{W}_{\mathbf{X}}$. (2) \textbf{Semantical Weighting:} For each clip and each query, we calculate its frame-to-frame similarity $\mathbf{W}_{\mathbf{C}_i}$ and frame-to-query similarity $\mathbf{W}_{\mathbf{C}_i,q}$ using cosine similarity between all frames $x_j \in \mathbf{C}_i$ and all query tokens $t_i\in {q}$. Frames and queries with low similarity are suppressed by their corresponding weights during optimization. (3) \textbf{Clip-wise Optimization:} Perturbation are optimized as per clip before outputting the final $\delta_{C_i}$ for adversarial video $\mathbf{X'}$.}
\end{figure*}

\begin{figure}[!t]
\centering
  \includegraphics[width=0.9\linewidth]{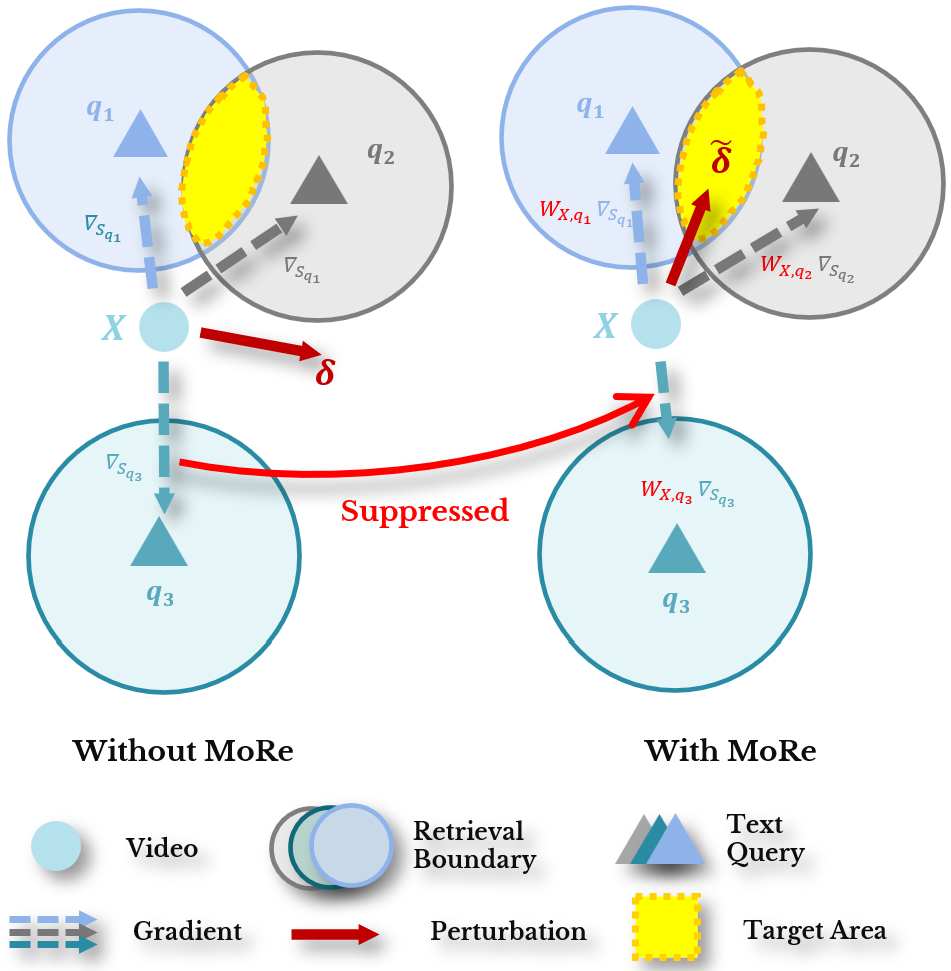}
\caption{\label{fig:qv}
An illustration of the effectiveness of MoRe in guiding optimization. Due to the conflicting gradients, $\delta$ in the left figure leads to suboptimal results without entering any boundary. After applying MoRe, the anomalous query ($q_3$) is suppressed, \textbf{yielding a more focused perturbation $\tilde\delta$ to push videos towards the target area} (yellow shape).
}
\end{figure}

\begin{algorithm}[!t]
\caption{Pseudo-code for ViPro. \textbf{MoRe} will be enabled for generating black-box attacks and disabled otherwise.}
\algrenewcommand\algorithmicrequire{\textbf{\textcolor{black}{Input:}}}
\algrenewcommand\algorithmicensure{\textbf{\textcolor{black}{Output:}}}
\label{alg}
\begin{algorithmic}[1]
\Require Video $\mathbf{X}$, target queries $\mathbf{Q}$, visual encoder $\mathcal{V}$, textual encoder $\mathcal{T}$, maximum PGD steps $K$, PGD step size $\alpha$, perturbation bound $\epsilon$
\Ensure Adversarial Video $\mathbf{X}'$
\tikzmark{startmore1}
\If{MoRe}\Comment{Clip video when MoRe Enabled}
\State $\mathbb{C}=[\mathbf{C}_1,\mathbf{C}_2,...,\mathbf{C}_M]\gets$ \Call{TempClip}{$\mathbf{X}$}
\For{$m \gets 1$ to $M$}
    \State $\mathbf{C}'_m\gets$ \Call{Attack}{$\mathbf{C}_m,\mathbf{Q}$} \Comment{Clip-wise perturbation}
\EndFor
\State $\mathbf{X}' \gets$ \Call{Concat}{$\mathbf{C}'_1,\mathbf{C}'_2,...,\mathbf{C}'_M$} \Comment{Concat all clips}
\tikzmark{endmore1}
\HighlightBlockf{startmore1}{endmore1}
\Else
\State $\mathbf{X}'\gets$ \Call{Attack}{$\mathbf{X},\mathbf{Q}$}
\EndIf

\Function {Attack} {$\mathbb{X},\mathbf{Q}$} \Comment{$\mathbb{X}$: input video}
    \State $\mathbf{F}_{\mathbb{X}}\gets\mathcal{V}(\mathbb{X}), \mathbf{F}_{\mathbf{Q}}\gets\mathcal{T}(\mathbf{Q})$ \Comment{Get features}
    \tikzmark{startmore2}
    \If{MoRe} \Comment{Get weights when MoRe enabled}
    \State $\mathbf{W}_{\mathbb{X}}\gets$\Call{CosSim}{$\mathbf{F}_{\mathbb{X}}$,$\mathbf{F}_{\mathbb{X}}$} \Comment{Temporal sim}
    \State $\mathbf{W}_{\mathbb{X},\mathbf{Q}}\gets$\Call{CosSim}{$\mathbf{F}_{\mathbb{X}}, \mathbf{F}_{\mathbf{Q}}$} \Comment{Frame-to-token sim}
    \EndIf\tikzmark{endmore2}
    \State Initialize ${\mathbb{X}}'_{0}\gets\mathbb{X}, \delta_{\mathbb{X}'_0}\gets0$
    \HighlightBlocks{startmore2}{endmore2}
    \For{$k \gets 1$ to $K$}
        \State ${\mathbb{X}'_k}\gets{\mathbb{X}}'_{k-1}+\delta_{{\mathbb{X}}'_{k-1}}$
        \State $\mathbf{F}_{{\mathbb{X}}'_{k}}\gets\mathcal{V}({\mathbb{X}'_{k}})$
        \State $\mathbf{S}\gets$ {$Sim\big(\mathbf{F}_{\mathbf{Q}},\mathbf{F}_{{\mathbb{X}}'_{k}}\big)$}   \Comment{Get sim using Eq.\ref{eq:sim}}
        \tikzmark{starmore3}
        \If{MoRe} \Comment{Calculate $\mathcal{L}_{more}$ using Eq.\ref{eq:l_more}}
        \State $\mathcal{L}\gets\mathcal{L}_{exp}\Big({W_{\mathbb{X}}}^{\textrm{\textbf{T}}}\cdot \Big|\mathbf{W}_{\mathbb{X},{\mathbf{Q}}}\odot\mathbf{S}\Big|\Big)$
        \tikzmark{endmore3}
        \HighlightBlockt{startmore3}{endmore3}
        \Else
        \State $\mathcal{L}\gets\mathcal{L}_{exp}(\mathbf{S})$ \Comment{Calculate $\mathcal{L}_{exp}$ in Eq.\ref{eq:lexp}}
        \EndIf
        \State $\delta_{{\mathbb{X}}'_{k-1}}\gets$ \Call{Proj}{$\alpha \nabla \mathcal{L},\epsilon$}   \Comment{PGD}
    \EndFor
    \State
    \Return $\mathbb{X}'_k$
    
\EndFunction
\end{algorithmic}
\end{algorithm}

\subsubsection{\textcolor{black}{Modality Refinement (MoRe)}}
\label{sec:more}
An overview of MoRe is presented in Figure.~\ref{fig:pipline}. For a chosen video $\mathbf{X}=[x_1,x_2,...,x_T]$ with $T$ frames and the target queries $\mathbf{Q} = [q_1,q_2,...,q_N]$, MoRe handles the finer-grained interaction between intra-modality and inter-modality in a temporal (intra) and a semantical (inter) respective. Specifically, we first conduct \textbf{temporal clipping} to group temporally related frames, and then perform \textbf{semantical weighting} to guide optimization using frame-to-token similarity for each query. In this way, MoRe will aid the optimization by guiding it towards target queries for enhanced transferability. Finally, each clip will be optimized separately and concatenated as the manipulated video ${\mathbf{X}}'$.

\noindent\textbf{Temporal Clipping.} To accommodate temporal abruptness among video frames (intra-modality), perturbation using averaged gradients from all frames could hinder the convergence of losses and lead to sub-par transferability. An intuitive solution is to perform per-frame optimization, but this would significantly increase computational complexity and break temporal information. Thus, we propose Temporal Clipping to identify temporally outlying frames and group frames into clips. Specifically, for each input video $\mathbf{X}$, we first calculate its frame-wise cosine similarity $\mathbf{W}_{\mathbf{X}}\in\mathbb{R}^{T\times T}$ as follows:
\begin{equation}
\label{eq:f2f}
\mathbf{W}_{\mathbf{X}}(i,j)=CosSim(x_i, x_j)
\end{equation}

We then calculate the similarity difference $\Delta_{\mathbf{W}_{\mathbf{X}}}$ as the temporal shifts between frames, where a larger value indicates an outlying frame with palpable temporal abruptness. Subsequently, we loop through all frames and clip them when exceeding the threshold $\gamma$. (See detailed pseudo-code in the Supplementary.) Temporal Clipping will output the clipped videos $\mathbb{C}=[\mathbf{C}_1,...,\mathbf{C}_M]$ for subsequent clip-to-query alignment.

\noindent\textbf{Semantical Weighting.} Similarly, the cross-modality alignment between clips and queries may vary significantly because of the semantic difference. Pushing videos towards them simultaneously may also lead to a `looser' convergence, i.e., failing to enter the retrieval boundary due to conflicting gradient, as shown in the left graph of Figure.\ref{fig:qv}. To avoid this, given a query $q$ and a video clip $\mathbf{C}$, we calculate the frame-to-query weights $\mathbf{W}_{\mathbf{C},q}$ using the mean of frame-to-token cosine similarity:
\begin{equation}
\label{eq:f2t}
 \mathbf{W}_{\mathbf{C},q} = \frac{\sum^N_{j}CosSim(x_i,t_j)}{N}, \forall x_i \in \mathbf{C}
\end{equation}
$\mathbf{W}_{\mathbf{C},q}$ is subsequently used for weighting the corresponding frames and query for optimal perturbations. In essence, semantic weighting suppresses anomalous queries during optimization to enable a `tighter' convergence towards other queries, as shown in the right graph of Figure.\ref{fig:qv}. The clip-wise temporal similarity $\mathbf{W}_{\mathbf{C}}$ is similarly used as intra-modality semantical weighting to guide frame-level optimization, resolving potential conflicts among gradients from different frames.

\noindent{\textbf{Clip-wise Perturbation.}}
Lastly, we optimize the video in a clip-wise manner and present the overall learning objectives of ViPro with MoRe as follows:
\begin{equation}
\label{eq:l_more}
   \mathcal{L}_{more}= \mathcal{L}_{exp}\Big({W_{\mathbf{C}}}_i^{\textrm{\textbf{T}}}\cdot \Big|\mathbf{W}_{\mathbf{C}_{i},{q}}\odot\mathbf{S}\Big|\Big), \forall \mathbf{C}_i \in \mathbb{C}
\end{equation}
where ${W_{\mathbf{C}_i}},\mathbf{W}_{\mathbf{C}_{i}{q}}, \mathbf{S}$ are all matrices with dimensions of $1\times T_{\mathbf{C}_i}$, $T_{\mathbf{C}_i}\times N$ and $T_{\mathbf{C}_i}\times N$. $T_{\mathbf{C}_i}$ is the number of frames within the clip $\mathbf{C}_i$. 

In summary, we incorporate MoRe for a finer-grained alignment of intra-/inter-modalities. Temporal Clipping helps identify and clip temporally outlying frames, while query weighting enables a guided optimization towards target queries by suppressing conflicting gradients. The two modules work cooperatively to push videos further towards target queries for enhanced transferability, which is comprehensively validated through experiments in Section.\ref{sec:as}.

Overall, ViPro is used for \textbf{white-/grey-box attacks}, while MoRe is enabled to generate transferable \textbf{black-box} attacks. We provide a pseudo-code for ViPro in Algorithm.\ref{alg}.

\section{Experiments}
\label{sec:exp}

\subsection{Experimental Settings}
\label{sec:ES}
\noindent \textbf{Datasets.} 
We test our attacks on the three prevailing datasets: MSR-VTT-1K (MSR-VTT)\cite{ref:MSRVTT} contains a 1K testset of YouTube videos, DiDeMo\cite{ref:DDM} contains a testset of 1065 Flickr videos, and ActivityNet \cite{ref:ANT} contains a 4.9K testset sampled from YouTube. To ensure equal testset sizes, we randomly sample a 1K subset for all datasets. We use MSR-VTT for model-wise evaluations.

\noindent \textbf{User Queries.} 
As mentioned in Table.\ref{tabl_as}, attackers could access \textbf{captions} for white-/grey-box settings and \textbf{categories} for the black-box setting. The acquisition of user queries is identical for all scenarios: assuming that all videos come with captions (categories) on T2VR platforms such as YouTube, attackers first use the captions of their own video to query the victim model for the top-$K$ retrieved videos and the corresponding captions (categories). These $K$ captions (categories) are regarded as the relevant queries of this video.

In our experiments, we use $K=20$ queries and randomly divide them evenly into train/test sets for the harshest scenario where users use {different similar but different queries than the ones used for training attacks}. \textit{All captions/categories used in our experiments are original from the datasets}. See details of data construction in the Supplementary.

\noindent\textbf{Victim Models.} 
We choose open-sourced leading T2VR models for MSR-VTT on PapersWithCode, including Singularity-17M (Sing)\cite{ref:singularity}, DRL-32B (DRL) \cite{DRL}, and Cap4Video-32B (C4V)\cite{C4V}. \textcolor{black}{The three models are selected as they representatively cover the general implementations of the `two-plus-one' paradigm of T2VR models, i.e., cross-encoder (Sing), MLP layers (DRL) and extra modules (C4V).} DRL and C4V are manually trained on MSR-VTT using the released code. All models take an input size of $3\times224\times224$. As for input frames, Sing uses 4 frames per video, while DRL and C4V take 12 frames. \textcolor{black}{We attack all frames of the video, and thus all frames  selected by the victim models are perturbed.}

For unimodal encoders, Sing uses BEiT \cite{ref:Beit} pre-trained on ImageNet-21K \cite{ref:imagenet} for the vision encoder and the first 9 layers of BERT\cite{ref:Bert} model for the text encoder. DRL and C4V use a Bert-based textual encoder and ViT as its visual encoder.
For cross-modal interaction, Sing uses a cross-encoder, and DRL uses a weight-based token interaction for cross-modality interaction. C4V uses a similar token interaction as DRL and introduces a sequence transformer as the video aggregator.

\noindent\textbf{Baselines.}
Due to the lack of directly comparable baselines, we use Co-Attack\cite{Coattack} and SGA\cite{SGA}, which are targeted at text-to-image retrieval (T2IR), as two baselines for comparison. (Attacks introduced by \citet{VTRA} are not included due to the lack of reproducible code.) We then apply the necessary and minimal adaptation to ensure that the two baselines comply with our settings. 

For both baselines, textual attacks are not implemented since attackers can only manipulate their updated videos and cannot interfere with user queries. And we reverse the original attacking objectives from \textbf{suppressing} ranks to \textbf{promoting} ranks. For SGA and ViPro, we use the white-box text-to-video similarities and cosine similarity for white-box and grey-box attacks, respectively. For white-box Co-Attack, we apply its strongest multimodal embedding setting, i.e., minimizing the distance between the perturbed videos and the multimodal embedding of the video-text pairs, denoted as \texttt{$\texttt{EMB}_{\texttt{Mul}}$}. For C4V/DRL, which do not feature multimodality embeddings, we modify the objectives to maximize the white-box text-to-video similarity, denoted as $\texttt{SIM}_{\texttt{T2V}}$. For grey-box Co-attack, we adopt the same cosine similarity for all models.

\noindent\textbf{Evaluation Metrics.}
We use the difference between vanilla and attacked R@1/5 as evaluation metrics for attacking performance, denoted as $\Delta R@1/5$. Results are presented in the average over the 1K testset for all models/datasets. Higher values are preferable as they indicate efficient attacks. 

\begin{table}[t!]
     \caption{\label{tabl_vipro_wb_d} White-box results on all datasets (ActivityNet/DiDeMo/MSR-VTT). The model is Singularity. Changes in R@K are presented in the parentheses. MoRe \textbf{disabled}.}
    \centering
   \resizebox{\linewidth}{!}{
   \begin{tabular}{c|c|c|c|c}
    \toprule
    {\textbf{Dataset}} &    {\textbf{Method}}    &  \textbf{R@1}(\%) $\uparrow$  & \textbf{R@5}(\%) $\uparrow$ &        \textbf{Average}(\%) $\uparrow$              \\
     \cmidrule{1-5}
    \multirow{5}*{ActivityNet\cite{ref:ANT}}  &   - &   4.78      &   24.44      &        14.61             \\
     \cmidrule{2-5}
                                &    Co-Attack\cite{Coattack} (\texttt{$\texttt{EMB}_{\texttt{Mul}}$})  &     3.96 (-0.82)      &   26.34 (+1.90)      &      15.15 (+0.54)              \\
                                &       SGA\cite{SGA}    &   7.99 (+3.20)      &   38.44 (+14.00)      &       23.22 (+8.61)               \\
                                &     ViPro  &    \textbf{46.66 (+41.88)}      &  \textbf{76.25 (+51.81)}      &        \textbf{61.46 (+46.85)    }        \\
                                  \cmidrule{2-5}
                   &     ViPro w/o $\epsilon$  &   {76.22 (+71.44)}      &   {90.77 (+66.33)}      &        {83.50 (+ 68.89)    }        \\

     \cmidrule{1-5}
     \multirow{5}*{DiDeMo\cite{ref:DDM}}  &   - &   4.90      &   24.71      &        14.78             \\
     \cmidrule{2-5}
                                &    Co-Attack\cite{Coattack} (\texttt{$\texttt{EMB}_{\texttt{Mul}}$})  &     5.42 (+0.52)      &   28.09 (+3.38)      &     16.76 (+1.95)              \\
                                &       SGA\cite{SGA}    &   9.58 (+4.68)      &   35.08 (+10.37)      &       22.33 (+7.53)               \\
                                &    ViPro    &     \textbf{40.54 (+35.64)}      &  \textbf{69.56 (+44.85)}      &        \textbf{55.03 (+40.25)    }        \\
                                \cmidrule{2-5}
          &     ViPro w/o $\epsilon$  &    {66.63 (+61.73)}      &   {84.90 (+60.19)}      &        {75.74 (+60.96)    }        \\

    \cmidrule{1-5}
    \multirow{5}*{MSR-VTT\cite{ref:MSRVTT}}  &   - &   4.87      &   22.94      &        13.91             \\
     \cmidrule{2-5}
                                &    Co-Attack\cite{Coattack} (\texttt{$\texttt{EMB}_{\texttt{Mul}}$})  &    9.51 (+4.64)      &   25.12 (+2.18)      &     17.32 (+3.41)              \\
                                &       SGA\cite{SGA}    & 19.30 (+14.43)      &   46.77 (+23.83)      &       33.04 (+19.13)               \\
                                &    ViPro   &     \textbf{47.81 (+42.94)}      &  \textbf{73.66 (+50.72)}      &        \textbf{60.74 (+46.83)    }        \\
                                  \cmidrule{2-5}
                   &     ViPro w/o $\epsilon$  &   {66.85 (+61.98)}      &   {82.53 (+59.59)}      &        {74.69 (+60.79)    }        \\

    \bottomrule
    \end{tabular}}

\end{table}

\begin{table}[t!]
     \caption{\label{tabl_vipro_wb_m} White-box results on all models (Sing/DRL/C4V) using MSR-VTT. Changes in R@K are presented in the parentheses. MoRe \textbf{disabled}. }
    \centering
   \resizebox{\linewidth}{!}{
   \begin{tabular}{c|c|c|c|c}
    \toprule
    {\textbf{Models}} &    {\textbf{Method}}    &  \textbf{R@1}(\%) $\uparrow$  & \textbf{R@5}(\%) $\uparrow$ &        \textbf{Average}(\%) $\uparrow$              \\
     \cmidrule{1-5}
     \multirow{5}*{Sing\cite{ref:singularity}} &   No Attack  &   4.87      &   22.94      &        13.91             \\
     \cmidrule{2-5}
                                &    Co-Attack\cite{Coattack}  (\texttt{$\texttt{EMB}_{\texttt{Mul}}$})  &     9.51 (+4.64)      &   25.12 (+2.18)      &     17.32 (+3.41)              \\ 
                                &       SGA\cite{SGA}&   19.30 (+14.43)      &   46.77 (+23.83)      &       33.04 (+19.13)               \\
                                &     ViPro   &    \textbf{47.81 (+42.94)}      &  \textbf{73.66 (+50.72)}      &        \textbf{60.74 (+46.83)    }        \\
                          \cmidrule{2-5}
           &     ViPro w/o $\epsilon$  &   {66.85 (+61.98)}      &   {82.53 (+59.59)}      &        {74.69 (+60.79)    }        \\

     \cmidrule{1-5}
     \multirow{5}*{DRL\cite{DRL}}  &   No Attack  &   4.29      &   15.76     &      10.03             \\
     \cmidrule{2-5}
                                &    Co-Attack\cite{Coattack} ($\texttt{SIM}_{\texttt{T2V}}$)  &     61.07 (+56.78)      &   83.89 (+68.13)      &     72.48 (+62.46)              \\
                                &       SGA\cite{SGA}   &   53.53 (+49.24)      &  80.37 (+64.61)      &       66.95 (+56.93)               \\
                                &     ViPro  &    \textbf{61.24 (+56.95)}      &  \textbf{84.62 (+68.86)}      &        \textbf{72.93 (+62.91)    }        \\
      \cmidrule{2-5}
          &     ViPro w/o $\epsilon$  &    {87.37 (+83.08)}      &   {97.09 (+81.33)}      &        {92.23 (+82.21)    }        \\

    \cmidrule{1-5}
     \multirow{5}*{C4V\cite{C4V}}  &   No Attack  &   3.76     &   13.42      &       8.59             \\
     \cmidrule{2-5}
                                &    Co-Attack\cite{Coattack} ($\texttt{SIM}_{\texttt{T2V}}$) &    75.04 (+71.28)      &   88.57 (+75.15)      &     81.81 (+73.22)              \\
                                &       SGA\cite{SGA}  &   72.13 (+68.37)      &   86.99 (+73.57)      &       79.56 (+70.97)               \\
                                &     ViPro  &    \textbf{75.20 (+71.44)}      &  \textbf{88.91 (+75.49)}      &        \textbf{82.06(+73.47)    }        \\
                          \cmidrule{2-5}
           &     ViPro w/o $\epsilon$  &    {93.61 (+89.85)}      &   {98.56 (+85.14)}      &        {96.09 (+87.50)   }        \\

    \bottomrule
    \end{tabular}}

\end{table}

\noindent \textbf{Hyperparameters.} 
For all attacks, we adopt PGD \cite{pgd} using $l_{\inf}$ norm with $\epsilon = \frac{16}{255}$, $\eta=2^7$, and the step size $\alpha = {1}/{255}$. Ablation studies for hyperparameters are given in Section.\ref{sec:as}. The random seed is fixed to 42. For the temporal clipping threshold $\gamma$, we empirically use $1-\frac{\sqrt3}{2}$ to indicate outlying frames, i.e., frame vectors deviating from the previous one by $\theta\geq\pi/6$. \textcolor{black}{We also provide an ablation on $\gamma$ in Section.\ref{sec:as}}. 

\subsection{Main Results}
We first provide white-box results to demonstrate the effectiveness of ViPro as an approximate upper bound. Afterwards, we present grey-box results to show the generalizability of ViPro under restricted knowledge. Finally, we conduct cross-model attacks for the transferability of all attacks.

\begin{table}[t!]
     \caption{\label{tabl_vipro_gb_d} Grey-box results on all datasets (ActivityNet/DiDeMo/MSR-VTT). The model is Sing. Changes in R@K are presented in the parentheses. MoRe \textbf{disabled}.}
    \centering
   \resizebox{\linewidth}{!}{
   \begin{tabular}{c|c|c|c|c}
    \toprule
    {\textbf{Dataset}} &    {\textbf{Method}}    &  \textbf{R@1}(\%) $\uparrow$  & \textbf{R@5}(\%) $\uparrow$ &        \textbf{Average}(\%) $\uparrow$              \\
     \cmidrule{1-5}
      \multirow{5}*{ActivityNet\cite{ref:ANT}}  &   No Attack  &   4.78      &   24.44      &        14.61             \\
     \cmidrule{2-5}
                                &    Co-Attack\cite{Coattack}  &     4.66 (-0.12)      &   25.52 (+1.90)      &      15.09 (+0.48)              \\
                                &       SGA\cite{SGA}   &   0.81 (-3.97)      &   8.24 (-16.20)      &       4.53 (-10.09)               \\
                                &     ViPro &    \textbf{16.19 (+11.41)}      &  \textbf{57.84 (+33.40)}      &        \textbf{37.02 (+22.41)    }        \\

     \cmidrule{1-5}
      \multirow{5}*{DiDeMo\cite{ref:DDM}}  &   No Attack  &   4.90      &   24.71      &        14.78             \\
     \cmidrule{2-5}
                                &    Co-Attack\cite{Coattack}  &    4.48 (-0.42)      &   22.49 (-2.22)      &     13.49 (-1.32)              \\
                                &       SGA\cite{SGA} &     1.23 (-3.67)      &   7.54 (-17.17)      &     4.39 (-10.42)            \\
                                &     ViPro  &    \textbf{8.92 (+4.02)}      &  \textbf{38.15 (+13.44)}      &        \textbf{23.51 (+8.73)    }        \\

    \cmidrule{1-5}
    \multirow{5}*{MSR-VTT\cite{ref:MSRVTT}}  &   No Attack  &   4.87      &   22.94      &        13.91             \\
     \cmidrule{2-5}
                                &    Co-Attack\cite{Coattack} &    4.66 (-0.21)      &   20.86 (-2.08)      &     12.76 (-1.15)              \\
                                &       SGA\cite{SGA}  &   0.98 (-3.98)      &   4.05 (-18.89)      &       2.52 (-11.39)               \\
                                &     ViPro  &    \textbf{8.71 (+3.84)}      &  \textbf{28.9 (+5.85)}      &        \textbf{18.76 (+4.85)    }        \\
              
    \bottomrule
    \end{tabular}}

\end{table}

\begin{table}[t!]
     \caption{\label{tabl_vipro_gb_m} Grey-box results on all models (Sing/DRL/C4V) using MSR-VTT. Changes in R@K are presented in the parentheses. MoRe \textbf{disabled}.}
    \centering
   \resizebox{\linewidth}{!}{
   \begin{tabular}{c|c|c|c|c}
    \toprule
    {\textbf{Models}} &    {\textbf{Method}}    &  \textbf{R@1}(\%) $\uparrow$  & \textbf{R@5}(\%) $\uparrow$ &        \textbf{Average}(\%) $\uparrow$              \\
     \cmidrule{1-5}
     \multirow{4}*{Sing\cite{ref:singularity}} &  No Attack &   4.87      &   22.94      &        13.91             \\
     \cmidrule{2-5}
                             &    Co-Attack\cite{Coattack}    &    4.66 (-0.21)      &   20.86 (-2.08)      &     12.76 (-1.15)              \\
                                &       SGA\cite{SGA}  &   0.98 (-3.98)      &   4.05 (-18.89)      &       2.52 (-11.39)               \\
                                &     ViPro &    \textbf{8.02 (+3.15)}      &  \textbf{28.31 (+5.37)}      &        \textbf{18.17 (+4.26)    }        \\

     \cmidrule{1-5}
      \multirow{4}*{DRL\cite{DRL}}  &  No Attack  &   4.29      &   15.76     &      10.03             \\
     \cmidrule{2-5}
                                &    Co-Attack\cite{Coattack}   &     38.61 (+34.328)      &   76.09 (+60.33)      &    57.35 (+47.33)              \\
                                &       SGA\cite{SGA}  &   39.41 (+35.12)      &  76.46 (+60.70)      &       57.97 (+47.91)               \\
                                &     ViPro &    \textbf{42.35 (+38.06)}      &  \textbf{78.39 (+62.63)}      &        \textbf{60.37 (+50.35)    }        \\

    \cmidrule{1-5}
      \multirow{4}*{C4V\cite{C4V}}  &   No Attack  &   3.76     &   13.42      &       8.59             \\
     \cmidrule{2-5}
                                &    Co-Attack\cite{Coattack}  &    0.93 (-2.83)      &   5.07 (-8.35)      &     3.00 (-5.59)              \\
                                &       SGA\cite{SGA}  &   1.28 (-2.48)      &   7.69 (-5.73)      &       4.49 (-4.11)               \\
                                &     ViPro &    \textbf{1.99 (-1.77)}      &  \textbf{10.53 (-2.90)}      &        \textbf{6.26 (-2.34)    }        \\

    \bottomrule
    \end{tabular}}
\end{table}

\noindent \textbf{White-Box.}
We present white-box results on all datasets in Table.\ref{tabl_vipro_wb_d} and all models in Table.\ref{tabl_vipro_wb_m} to demonstrate the effectiveness of ViPro. Overall, all white-box attacks significantly promote video ranks to much higher R@1/5 for all datasets/models, showing the impact of our proposed attacking paradigm across datasets and models. Among them, ViPro achieves the best results over Co-Attack and SGA for all datasets and all models. For dataset-wise comparison, ViPro has a consistent lead for $30\%$ on R@1 over all datasets, especially on ActivityNet, which surpasses Co-Attack and SGA by $\sim$43/38\%, respectively. ViPro also achieves SOTA performance for model-wise comparison. 

For Co-Attack, it constantly performs the worst on all datasets on the Sing model because it applies the KL divergence to optimize using the implicit multimodal embedding to promote videos, which is inefficient in guiding videos towards queries. Once adapted to white-box similarity on DRL and C4V, Co-Attack's performance benefits significantly. As for SGA, although it directly utilizes the white-box similarity, it adopts the $\mathcal{L}_{neg}$ defined in Eq.\ref{eq:lnaive} plus an aggressive augmentation that includes 4 augmented images. The former leads to suboptimal results since $\mathcal{L}_{neg}$ treats all targets identically, while the latter harms the optimization by augmenting semantically irrelevant videos into training. Specifically, such augmentation could be more damaging on `sensitive' models such as Sing, which is expected to have smaller retrieval boundaries, making it easier for augmented videos to corrupt the semantic correlations between videos and queries. In contrast, DRL and C4V are expected to be less `sensitive' as it has larger retrieval boundaries since SGA shows comparable performance on these models.

Lastly, we present results without perturbation bounds $\epsilon$ to demonstrate the experimental upper bound for ViPro attacks to validate our hypothesis on retrieval boundaries. Results on all datasets/models w/o $\epsilon$ show that, while gaining certain enhancements, ViPro cannot hit 100\% in R@1 and R@5, 
implying that the distribution of datasets and model embeddings determines the upper bound of ViPro attacks. We will further elaborate on this with an in-depth analysis in Section.\ref{sec:discussion}.

\begin{table*}[t!]
     \caption{\label{tabl_vipro_bb} Black-Box results of all attacks on MSR-VTT on all models. Results are evaluated using \textbf{categories}. MoRe \textbf{enabled}.}
    \centering
   \resizebox{\linewidth}{!}{
   \begin{tabular}{c|c|c|c|c|c|c|c|c}
    \toprule
 {\textbf{Source}} &  \multirow{2}*{\textbf{Method}} &   \multicolumn{2}{c|}{\textbf{Sing}} &    \multicolumn{2}{c|}{\textbf{DRL}}  &  \multicolumn{2}{c|}{\textbf{C4V} } &  \multirow{3}*{\textbf{Average}(\%)$\uparrow$} \\ 
 \cmidrule{3-8}
  {\textbf{Model}} &                                 & \textbf{R@1}(\%) $\uparrow$  & \textbf{R@5}(\%) $\uparrow$  & \textbf{R@1}(\%) $\uparrow$  & \textbf{R@5}(\%) $\uparrow$ &    \textbf{R@1}(\%) $\uparrow$  & \textbf{R@5}(\%) $\uparrow$   \\ 
     \cmidrule{2-8}
    \cmidrule(lr){1-9}
                                                 - &   No Attack  &    5.21  &    9.79    &   1.42    &      6.32      &   0.79      &   3.00      &      -       \\
     \cmidrule(lr){1-9}
    \multirow{3}*{Sing\cite{ref:singularity}}    &    Co-Attack  \cite{Coattack} &    --     &   --    &      0.63 (-0.79)      &  5.53 (-0.79)      &    0.16 (-0.63)      &   1.74 (-1.26)     &  2.02 (-0.87) \\
                                                 &    SGA  \cite{SGA}     &      --     &       --    &         0.70 (-0.72)      &  6.00 (-0.32)      &  \textbf{ 0.32 (-0.47)}      &   1.74 (-1.26)     &  2.19 (-0.69) \\
                                                &     ViPro  &    --     &  --   &       \textbf{1.58 (+0.16)}      &  \textbf{8.53 (+2.21)}      &     \textbf{0.32 (-0.47)}      &  \textbf{3.32 (+0.32)}     &    \textbf{3.44 (+0.56)}  \\
    \cmidrule(lr){1-9}
     \multirow{3}*{DRL\cite{DRL}}              &    Co-Attack \cite{Coattack}  &  \textbf{7.74 (+2.53)}   &   15.01 (+5.22)    &   --  &  --  &   17.69 (+16.90)     &   33.33 (+30.33)     &  18.44 (+13.75) \\
                                               &    SGA \cite{SGA}     &    7.58 (+2.37)    &   15.32 (+5.53)    &   --  &   --  &   22.59 (+21.80)     &   40.13 (+37.13)     &  21.41 (+16.71) \\
                                                &     ViPro &      \textbf{7.74 (+2.53)}    &    \textbf{16.11 (+6.32)}    &    --      &  --      &   \textbf{26.05 (+25.26)}      &    \textbf{45.31 (+42.31)}     &  \textbf{23.80 (+19.11)} \\
    \cmidrule(lr){1-9}
     \multirow{3}*{C4V\cite{C4V}}             &      Co-Attack \cite{Coattack}   &   5.69 (+0.48)   &   13.90 (+4.11)   &     5.21 (+3.79)      &  20.54 (+14.22)      &   --      & --     &  11.34 (+5.65) \\
                                             &      SGA \cite{SGA}        &   6.95 (+1.74)   &   13.74 (+3.95)   &      {7.11 (+5.69)}      &   {23.70 (+17.38)}      &   --      &   --     &  12.88 (+7.19) \\
                                         &     ViPro   &    \textbf{7.63 (+2.42)}   &     \textbf{13.99 (+4.20)}   &     \textbf{26.05 (+24.63)}      &  \textbf{46.31 (+39.99)}      &     --      &   --     &  \textbf{23.50 +(17.81)} \\
    \bottomrule
    \end{tabular}}

\end{table*}

\noindent\textbf{Grey-Box.}
We then present the results of all attacks under the grey-box setting for all datasets/models. In Table.\ref{tabl_vipro_gb_d} and Table.\ref{tabl_vipro_gb_m}, SGA and Co-attack fail to promote the R@1 of perturbed videos for most cases, incurring negative promotion after attacks. In contrast, our ViPro manages to generalize well and retain its advantages over other baselines for all datasets. For example, on DiDeMo and MSR-VTT, our ViPro is the only method that successfully promotes manipulated video, while other baselines are shown to be inefficient for grey-box attacks. For model-wise comparison, ViPro also achieves the SOTA results for all models. Specifically, SGA experienced large performance drops on all datasets, echoing our conclusion on the damage brought by augmentation on `sensitive' models. These leads further validate the effectiveness of our proposed ViPro over SGA and Co-Attack for promoting videos.

Compared to white-box settings, all grey-box attacks experience a performance downgrade. Specifically, we observe inconsistency in performance drops compared to their white-box counterparts. For example, attacks on DRL experienced the minimum degradations by only 15/9/12\% for Co-Attack/SGA/ViPro, respectively. On the other hand, attacks on C4V decrease significantly by over 60\% on average for all attacks, which is caused by the introduction of an extra video aggregator. \textcolor{black}{We provide an ablation to validate the impact of video aggregator in the Table.III of the Supplementary, where all attacks show significant performance boosts with access to the aggregator.} We will discuss these findings in Section.~\ref{sec:discussion}.

\noindent\textbf{Black-Box.}
We finally evaluate cross-model transferability in the black-box settings following the definition in Table.\ref{tabl_as}, where attackers can only access the categorical information from the video. Thus, black-box attackers could only target the annotated category to promote its rank under this category.

We use the categories provided in the original paper in \cite{ref:MSRVTT}. Specifically, we use the 10 out of 20 categories with the best zero-shot R-Precision on all models. Details on evaluations for all models and categories are provided in the Supplementary.

Results are given in Table.\ref{tabl_vipro_bb}. Overall, ViPro retains its lead over all baselines for all black-box models, leading Co-Attack and SGA by 6\% and 4\% on average. Specifically, when using C4V as the source model, ViPro significantly outperforms Co-Attack and SGA by 21/26\% and 19/23\% for R@1/5 when attacking DRL models. ViPro also surpasses the other two baselines by 5\% and 2\% on DRL-generated attacks, yielding a maximum 25.26\% promotion on R@1 and 42.31\% on R@5 when attacking C4V. As for the attacks optimized using Sing, our ViPro is the only method that achieves improved overall average R@1/5 among all methods, yielding a 0.5\% promotion over Co-Attack (-0.87\%) and SGA (-0.69\%). The results validate the advantages of ViPro + MoRe for transferability.

Besides, comparing attacks generated by different models, we find that attacks from Sing exhibit the least transferability than those generated by DRL and C4V, while DRL and C4V show better transferability to the others. This observation aligns with the consensus that models exhibit better/worse transferability on similar/dissimilar models. C4V and DRL models adopt the same ViT encoders, while Sing uses BEiT. \textcolor{black}{Adversarial perturbations optimized for Sing tend to overfit to its embedding and do not generalize to DRL/C4V, resulting in lowered R@k compared to the no-attack baselines. This phenomenon highlights the necessity of carefully designed attacks, such as our ViPro, to mitigate architectural mismatch.}

\subsection{Ablation Study}
\label{sec:as}
In this section, we first demonstrate the effectiveness of MoRe in improving transferability, with a specific comparison between temporal clipping and random clipping. Afterwards, we conduct a qualitative evaluation on the impact of the number of training targets on optimizing ViPro, yielding counter-intuitive yet inspiring results. We finally provide an ablation study on the hyperparameters used in PGD.

\begin{table}[t!]
     \caption{\label{tabl_as_d} Ablation studies on the effectiveness of MoRe to boost transferability. $\mathcal{R}$ denotes random clipping. Results are given as the average of each source model. MoRe achieves the \textbf{best overall black-box performances for ALL models}. MoRe \textbf{enabled} unless specified.}
    \centering
   \resizebox{\linewidth}{!}{
   \begin{tabular}{c|c|c|c|c}
    \toprule
    \textbf{Source Model}  & {\textbf{Method}}  &  \textbf{R@1}(\%) $\uparrow$  & \textbf{R@5}(\%) $\uparrow$ & \textbf{Average} (\%) $\uparrow$  \\ 
     \cmidrule{1-5}     
    \multirow{3}*{Sing\cite{ref:singularity}} &  ViPro w/o MoRe        &    {0.87(-0.24)} &  5.92 (+1.26) &  3.39 (+0.51) \\
                                              &   ViPro $\mathcal{R}$   &    0.79 (-0.32)      &   5.37 (+0.71)    &     3.08 (+0.20)      \\
                                              &   ViPro &     \textbf{0.95 (-0.16)}      &   \textbf{5.93 (+1.27)}   &      \textbf{3.44 (+0.56)}      \\
     \cmidrule{1-5}
 \multirow{3}*{DRL\cite{DRL}} &  ViPro  w/o MoRe     &    14.06 (+11.06) &  27.25 (+20.86) &  20.66 (+15.96) \\
                           &  ViPro $\mathcal{R}$   &    15.49 (+12.49)      &   27.57 (+21.17)    &   21.53 (+16.83)      \\
                          &   ViPro  &     \textbf{ 16.90 (+13.90)}      &   \textbf{30.71(+24.32)}   &      \textbf{23.80 (+19.11)}      \\
     \cmidrule{1-5}
 \multirow{3}*{C4V\cite{C4V}} &  ViPro w/o MoRe      &    15.33 (+12.01) & 25.68 (+17.62) & 20.50 (+14.82) \\
                               &  ViPro $\mathcal{R}$  &   12.00 (+8.69)      &  26.74 (+18.60)    &     19.37 (+13.69)      \\
                               &  ViPro &     \textbf{16.84 (+13.53)}      &   \textbf{30.15 (+22.10)}   &      \textbf{23.50 (+17.81)}      \\

    \bottomrule
    \end{tabular}}

\end{table}

\begin{table*}[t!]
     \caption{\label{tabl_wb_def} White-box defense evaluations using Temporal Shuffling and JPEG Compression on Sing, DRL, and C4V. Changes after attacks are presented in parentheses. \textbf{Our ViPro retains its superiority under both defences.} MoRe \textbf{disabled}.}
    \centering
   \resizebox{\linewidth}{!}{
   \begin{tabular}{c|c|c|c|c|c|c|c|c}
    \toprule
  \multirow{2}*{\textbf{Defense}} &  \multirow{2}*{\textbf{Attack}} &   \multicolumn{2}{c|}{\textbf{Sing}} &    \multicolumn{2}{c|}{\textbf{DRL}}  &  \multicolumn{2}{c|}{\textbf{C4V} } &  \multirow{3}*{\textbf{Average}(\%)$\uparrow$} \\ 
 \cmidrule{3-8}
   &                       & \textbf{R@1}(\%) $\uparrow$  & \textbf{R@5}(\%) $\uparrow$  & \textbf{R@1}(\%) $\uparrow$  & \textbf{R@5}(\%) $\uparrow$ &    \textbf{R@1}(\%) $\uparrow$  & \textbf{R@5}(\%) $\uparrow$   \\ 
     \cmidrule{2-8}
    \cmidrule(lr){1-9}
                                                 - &   No Attack  &    4.87   &    22.94    &   4.29    &     15.76     &   3.79      &  13.73      &      -       \\
     \cmidrule(lr){1-9}
    \multirow{3}*{ Temporal Shuffling\cite{ts}}    &    Co-Attack  \cite{Coattack} &    8.33 (+3.46)     &   21.48(-1.46)    &     38.82 (+34.53)      &   62.95 (+47.19)   &   \textbf{55.74 (+51.95)}     &  \textbf{73.58 (+59.85)}     &  43.48 (+32.59) \\
                               &    SGA  \cite{SGA}     &     17.11 (+12.24)    &      41.83 (+18.89)    &      36.05 (+31.76)      &   60.42 (+44.66)   &  53.20 (+49.41)    &   71.24 (+57.51)    &  46.64 (+35.75) \\
                                 &     ViPro   &   \textbf{38.39 (+33.52)}     &  \textbf{64.67 (+41.73)}   &       \textbf{39.32 (+35.03)}      &  \textbf{63.95 (+48.19)}      &    55.51 (+51.72)      &  73.28 (+59.55)     &    \textbf{55.85 (+44.96)}  \\
    \cmidrule(lr){1-9}
    \multirow{3}*{ JPEG Compression\cite{jpeg}}    &    Co-Attack  \cite{Coattack} &   8.49 (+3.62)     &   23.18(+0.24)    &     53.46 (+49.17)      &   77.55 (+61.79)   &   {67.55 (+63.76)}     &  {84.24 (+70.51)}     &  52.41 (+41.52) \\
                               &    SGA  \cite{SGA}     &    19.21 (+14.34)    &     46.59 (+23.65)   &      52.93 (+48.64)      &  77.80 (+62.04)  &  \textbf{67.82 (+64.03)}   &  84.20 (+70.47)    & 58.09 (+47.20) \\
              $(q=75)$                &     ViPro   &   \textbf{34.85 (+29.98)}     &  \textbf{63.67 (+40.73)}   &       \textbf{53.57 (+49.28)}      &  \textbf{78.00 (+62.24)}      &    67.76 (+66.97)      &  \textbf{84.26 (+70.53)}     &    \textbf{63.69 (+52.79)}  \\
 
    \bottomrule
    \end{tabular}}

\end{table*}

\begin{table*}[t!]
     \caption{\label{tabl_gb_def} Grey-Box defense evaluations using Temporal Shuffling and JPEG Compression on Sing, DRL, and C4V. Changes after attacks are presented in parentheses. \textbf{Our ViPro retains its superiority under both defences.} MoRe \textbf{disabled}.}
    \centering
   \resizebox{\linewidth}{!}{
   \begin{tabular}{c|c|c|c|c|c|c|c|c}
    \toprule
  \multirow{2}*{\textbf{Defense}} &  \multirow{2}*{\textbf{Attack}} &   \multicolumn{2}{c|}{\textbf{Sing}} &    \multicolumn{2}{c|}{\textbf{DRL}}  &  \multicolumn{2}{c|}{\textbf{C4V} } &  \multirow{3}*{\textbf{Average}(\%)$\uparrow$} \\ 
 \cmidrule{3-8}
   &                       & \textbf{R@1}(\%) $\uparrow$  & \textbf{R@5}(\%) $\uparrow$  & \textbf{R@1}(\%) $\uparrow$  & \textbf{R@5}(\%) $\uparrow$ &    \textbf{R@1}(\%) $\uparrow$  & \textbf{R@5}(\%) $\uparrow$   \\ 
     \cmidrule{2-8}
    \cmidrule(lr){1-9}
                                                 - &   No Attack  &    4.87   &    22.94    &   4.29    &     15.76     &   3.79      &  13.73      &      -       \\
     \cmidrule(lr){1-9}
         \multirow{3}*{Temporal Shuffling\cite{ts}}              &    Co-Attack \cite{Coattack}  &  4.65 (-0.22)   &  19.88 (-3.06)    &   21.08 (+16.79)  &  53.31 (+37.55)  &   0.47 (-3.32)     &   3.73 (-10.00)     &  17.19 (+6.29)\\
                                 &    SGA \cite{SGA}     &   0.89 (-3.98)    &   3.93 (-19.01)    &   24.56 (+20.27) &  60.58 (+44.82)  &   1.19 (-2.60)    &  7.32 (-6.41)     & 16.41 (+5.52) \\
                             &     ViPro   &      \textbf{8.44 (+3.57)}    &    \textbf{27.44 (+4.50)}    &    \textbf{37.84 (+33.55)}       &  \textbf{72.09 (+56.33)}       &   \textbf{1.83 (-1.96)}      &    \textbf{10.15 (-3.58)}     &  \textbf{26.30 (+15.40)}  \\
      \cmidrule(lr){1-9}
         \multirow{3}*{ JPEG Compression\cite{jpeg} }        &    Co-Attack \cite{Coattack}  &  5.06 (+0.19)   &  20.53(-2.41)    &   30.94 (+26.65)  &  67.22 (+51.46)  &   0.22 (-3.57)     &   1.92 (-11.81)     &  20.98 (+10.09)\\
                                           &    SGA \cite{SGA}     &  1.14 (-3.73)    &  4.52 (-18.42)    &  36.51 (+32.22) &  73.45 (+57.69)  &   1.27 (-2.52)    &  6.91 (-6.82)     &  20.63 (+9.74) \\
            $(q=75)$            &     ViPro   &      \textbf{7.80 (+2.93)}    &    \textbf{28.26 (+5.32)}    &    \textbf{46.26 (+41.97)}       &  \textbf{79.79 (+64.03)}       &   \textbf{1.57 (-2.22)}    &    \textbf{8.37 (-5.36)}     &  \textbf{28.68 (+17.78)}  \\
    \bottomrule
    \end{tabular}}

\end{table*}

\noindent\textbf{Modality Refinement.}
To validate the viability of Temporal Clipping, we provide results without MoRe and MoRe with random clipping (denoted as MoRe $\mathcal{R}$) for comparison. Results are presented as the average results of each source model for clarity, as shown in Table.\ref{tabl_as_d}. First of all, we observe sub-par or the worst performance for the random clipping, i.e., MoRe $\mathcal{R}$. For example, on the Sing model, random clipping leads to downgraded performance for both R@1 and R@5, yielding the worst results for the model. Moreover, the R@1 also drops significantly by over 3\% on attacks generated by C4V, implying that randomly clipping the video would potentially incur worse attacking performance. On the other hand, our ViPro + MoRe consistently achieves the best performance over its counterparts with random clipping and ViPro itself across models. For example, our MoRe brings a 2.9/3\% boost over ViPro on DRL regarding R@1/5, and a 1.5/5\% boost on C4V regarding R@1/5. For the most challenging source model, our ViPro + MoRe also boosts the R@1/5 considerably. This validates the effectiveness of our MoRe to 'tighten' the convergence of losses during the optimization of perturbations by clustering temporally similar frames and guiding optimization with query weighting.

\noindent\textbf{Number of Training Queries \textcolor{black}{(K)}.}
Intuitively, the number of training queries for optimizing ViPro could significantly influence attack performance. Intuition may be that ViPro would benefit from larger numbers of training queries as it acquires more knowledge about the victim datasets. To this end, we trial different numbers of training queries on all datasets and models, ranging from 10 (experiment setting) to 500 (half of the testset).

\begin{figure}[t!]
\centering
\includegraphics[width=\linewidth]{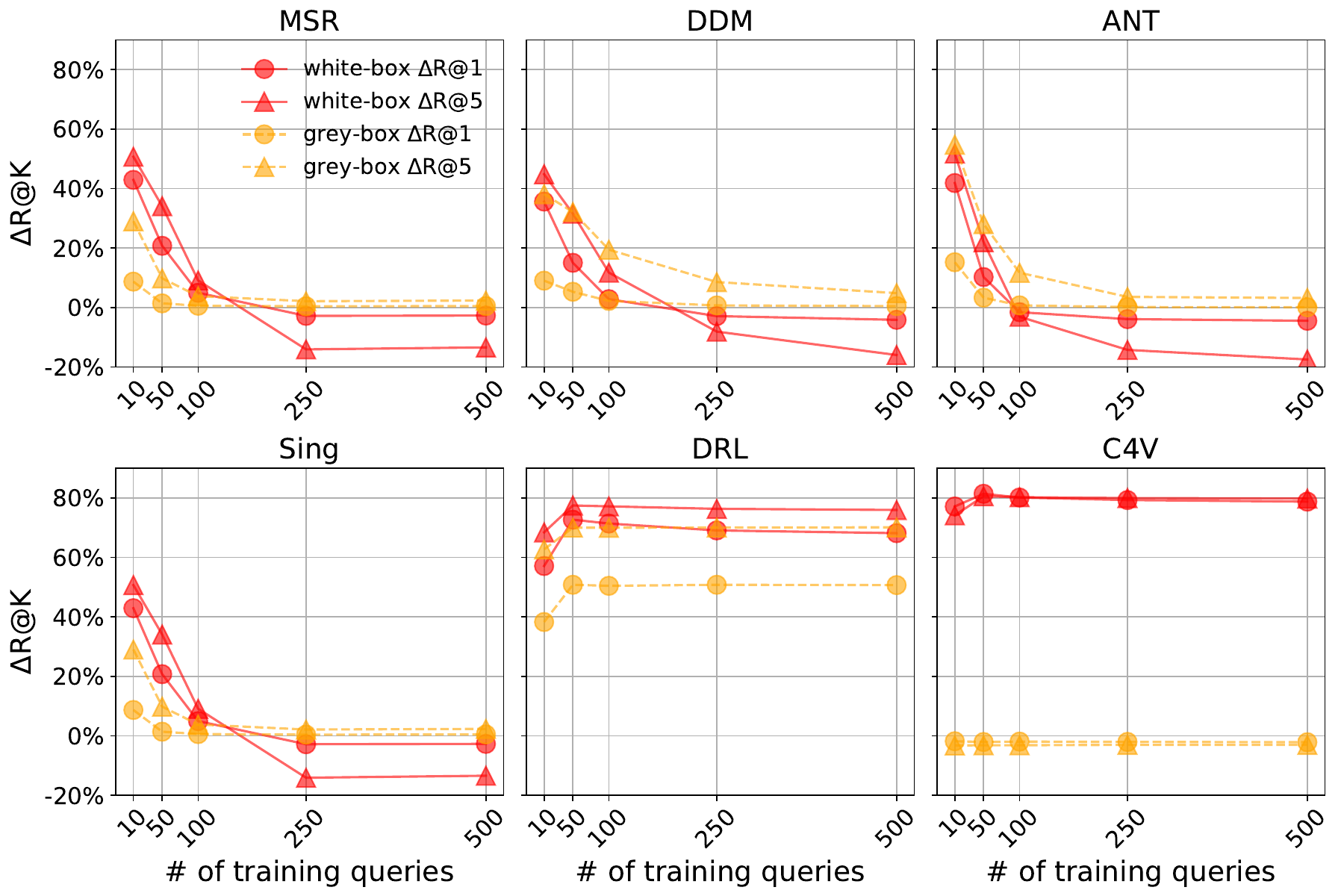}
  \caption{
The impact of varying numbers of training queries for white-box (solid lines) and grey-box (dashed lines) ViPro on all datasets (top row) and all models (bottom row). We present $\Delta$R@1 (red) and $
\Delta$R@5 (orange).  }
\label{fig:num_targets}
\end{figure}

As shown in Figure.\ref{fig:num_targets}, we find counter-intuitive results as the number of training queries increase for both white-\&grey-box attacks: for dataset-wise comparison (top row), the attacking performance suffers from a constant drop, yielding an over 40\% degradation as K reaches 500; for model-wise comparison (bottom row), the results on DRL and C4V exhibits a rise-to-plateau trend, with a minor drop in the late stage. Specifically, for DRL and C4V, using only 10 training queries can efficiently achieve more than 85\% of the max performance when using more queries, where as for Sing, using 10 already achieves the best results over others. \textcolor{black}{Interestingly, white-box attacks show subpar performance compared to grey-box attacks for Sing (i.e., top rows of Fig.\ref{fig:num_targets}) . This is because as the number of training queries increases, more irrelevant or less relevant queries are included in the training set, making the training queries `noisier' w.r.t. the test set. Consequently, both white-box and grey-box attacks exhibited a performance degradation for larger training sets ($K>10$), while white-box results experienced a severer drop as they are more susceptible to these `noises' than grey-box attacks because of the direct access to the model embedding. The reason why Sing is the only model exhibiting such a pattern originates from its model embedding, which will be discussed in Section.\ref{sec:discussion}.} Overall, we can draw the following conclusion: Using more queries does not necessarily boost attacking performance, while \textbf{a small number of queries is sufficient to generate effective attacks for both white-box and grey-box ViPro.}

\begin{figure}[t!]
\centering
\includegraphics[width=\linewidth]{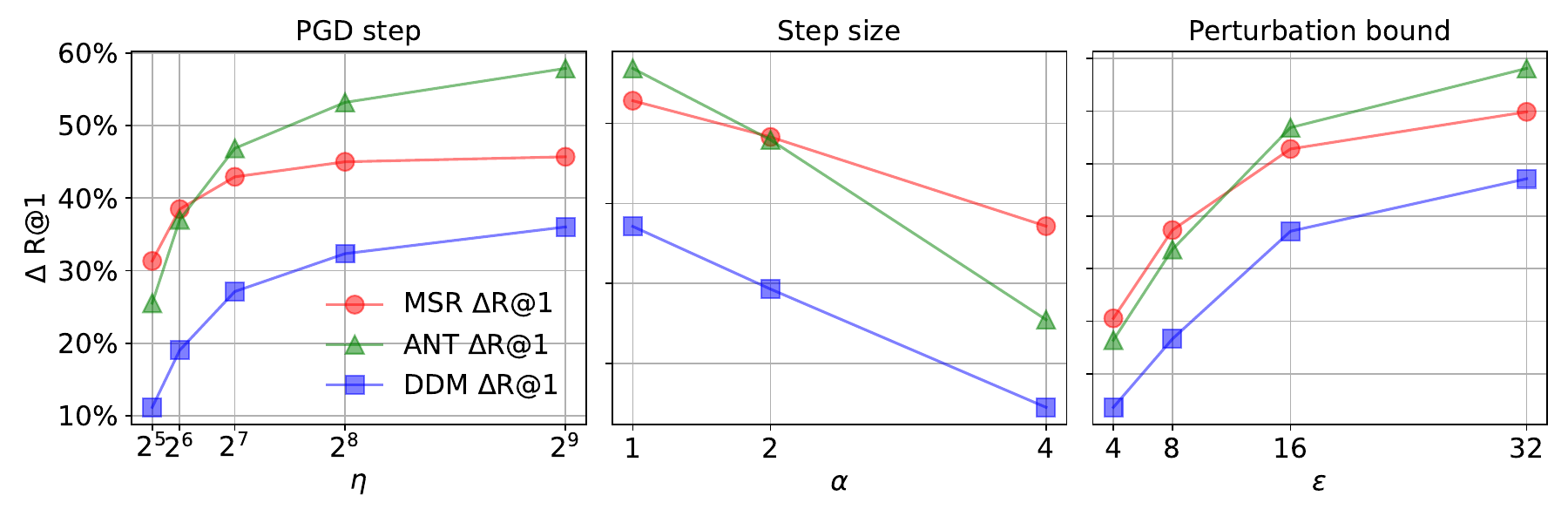}
\caption{Ablation studies on PGD hyperparameters, showing results for max PGD steps $\eta$, step size $\alpha$, and perturbation bound $\epsilon$, respectively. Results are presented in R@1 changes after attacks, denoted as $\Delta$R@1. R@5 results are not presented for better visibility.}
\label{fig:hyperpara}
\vspace{-0.5cm}
\end{figure}

\noindent\textbf{Hyperparameters.}
Ablations on PGD hyperparameters on all datasets are presented in Figure.\ref{fig:hyperpara}, showing the results for max steps $\eta$, step size $\alpha$, and perturbation bound $\epsilon$. We use white-box results on the Sing model for all datasets and present R@1 changes to demonstrate attacking performance.

For the max PGD step size $\eta$, we observe a monotonic increase as it increases from $2^4$ to $2^9$ for all datasets. Considering the computational cost of perturbing videos, we choose $2^7$ for balanced results between effectiveness and efficiency. Although further increasing $\eta$ would boost attacking performance, such boosts come at the cost of doubled or more training time. For the step size $\alpha$, we test values ranging from 1 to 4. Consistent with our previous speculation, ViPro suffers from a steady performance drop as the step sizes increase. This echoes with our illustration in Fig.\ref{fig:vl} that video promotion is an intricate attack that requires sophisticated optimization for optimal results. Finally, a similar trend as $\eta$ is found for the perturbation budget $\epsilon$. Thus, we empirically choose $\epsilon=16$ for better effectiveness while maintaining imperceptibility. We also use a user study to examine the stealthiness of our ViPro over Co-Attack and SGA under the same $\epsilon$ in Sec.\ref{sec:defense}.

\begin{figure}[t!]
\centering
\includegraphics[width=0.7\linewidth]{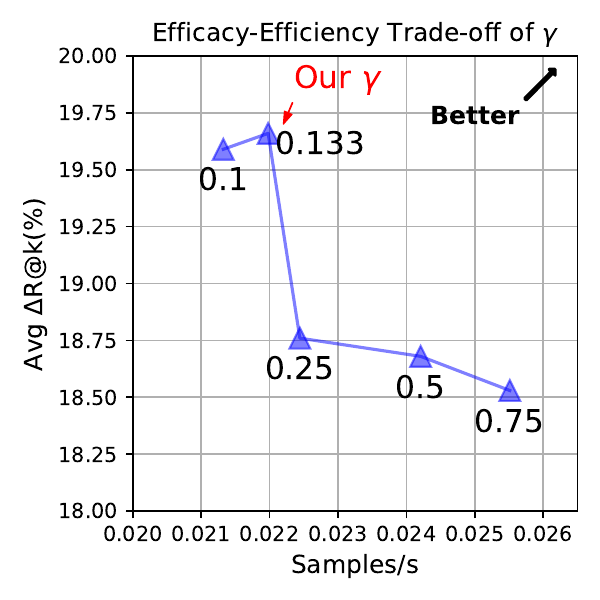}
\caption{Efficacy-Efficiency curve of various $\gamma$. Efficacy is presented using averaged $\Delta$R@1/5 of all models (denoted as Avg $\Delta$R@k), and efficiency is presented using the number of processed samples per second (denoted as samples/s). Overall, an evident trade-off can be observed: choosing smaller $\gamma$ (0.1/0.133) yields better efficacy at the cost of slower optimization, while larger $\gamma$ (0.5/0.75) leads to faster optimization but lower efficacy. We empirically chose $\gamma=0.133$ for best efficacy. }
\label{fig_gamma}
\vspace{-0.2cm}
\end{figure}

\textcolor{black}{Lastly, we investigate the impact of the clipping threshold $\gamma$. Overall, $\gamma$ requires reasonable selection because of the trade-off exists between efficiency (samples per second) and efficacy (ASR). To visualize such a trade-off, we plot an Efficiency-Efficacy evaluation for a set of $\gamma = 0.1, 0.133$ (our $\gamma$)$, 0.25, 0.5, 0.75$, as shown in Fig.\ref{fig_gamma}. Intuitively, choosing a larger $\gamma$ (0.5,0.75) would boost optimization efficiency, at the cost of downgraded transferability, while a smaller $\gamma$ (0.1,0.133) would yield better transferability but lower speed. We empirically chose $\gamma$ for the best $\Delta$R@k, i.e., $1-\cos(\pi/6)$, indicating angles between vectors no greater than $\pi/6$.}

\subsection{Evaluation under Realistic Scenarios}
\label{sec:defense}
In reality, attacks against T2VR may undergo defenses such as JPEG compression during upload. Besides, it is also vital that manipulated videos do not contain artifacts or visual clues that are obvious to human users. Consequently, to comprehensively evaluate the performance of all attacks under realistic scenarios, we perform evaluations for all attacks regarding both robustness against defences and human evaluations.

\begin{figure}[t!]
\centering
\includegraphics[width=1\linewidth]{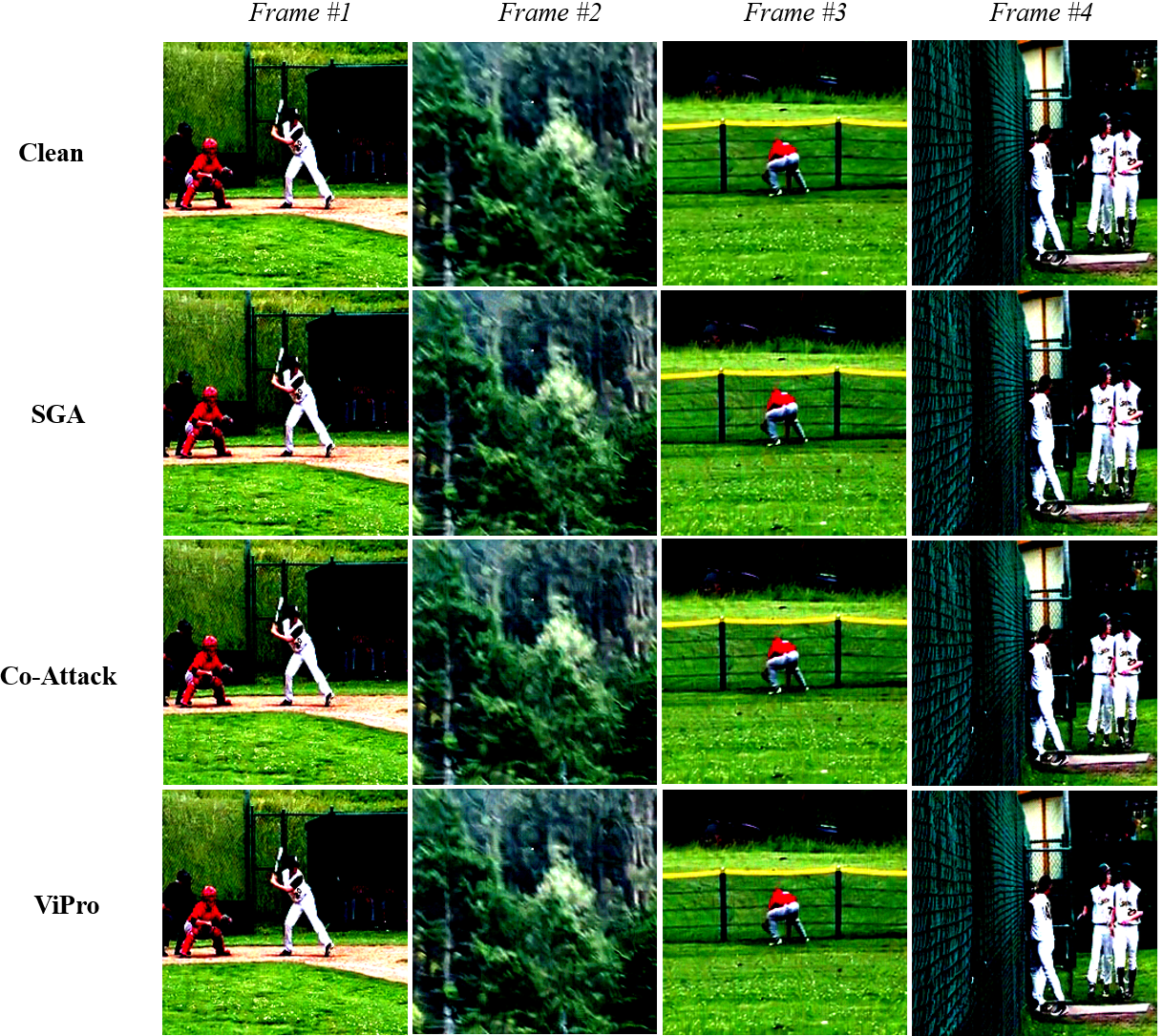}
  \caption{
Visualization of original frames and manipulated frames for all attacks (white-box). From top to bottom: \textbf{Clean, SGA, Co-Attack, ViPro (ours)}. SGA has the most obvious patterns and artifacts. Co-Attack has on-par stealthiness as our ViPro with minor visual clues, while \textbf{our ViPro remains stealthy with the best performance}. }
\label{fig:eg_frames}
\vspace{-0.5cm}
\end{figure}

\noindent\textbf{Robustness against Defences.}
To verify the robustness of all attacks under defence, we adopt JPEG Compression (JC) \cite{jpeg} and Temporal Shuffling (TS) \cite{ts}, two well-recognized image-based and video-based defences, to evaluate all included attacks on all models for white-box and grey-box scenarios. Following the settings in the original papers, we use JPEG with $q=75$, and for TS, we use $h_1=2, h_2=1$ for Sing (uses 4 frames), and $h_1=4, h_2=2$ for DRL and C4V (use 12 frames). Results are presented in Table.\ref{tabl_wb_def} and Table.\ref{tabl_gb_def}. Our ViPro shows the best overall robustness against both image-based and video-based defences, leading Co-Attack by 10/9\% and SGA by 7/9\% for white-/grey-box settings, respectively. 

\noindent\textcolor{black}{\textbf{Imperceptibility.}}
We finally provide evaluations on the visual imperceptibility of all attacks. We first present a visualization of all attacks in Fig.~\ref{fig:eg_frames}, from which we find SGA as the most suspicious compared to Co-Attack and ViPro. We attribute this to its more aggressive data augmentation strategy that leads to diverse but perceptible perturbations. 

\textcolor{black}{We start by providing numerical evaluation of all attacks using SSIM, PSNR, and LPIPS on all models, as presented in Table.\ref{imper_eval}. We find ViPro shows better or on-par imperceptibility with much better performance than Co-Attack and SGA.} We further conducted a user study with 43 shuffled groups of videos for each attack. Based on results from 17 human experts (i.e., equipped with background knowledge about adversarial attacks), ViPro is chosen as the stealthiest with \textbf{42.69\%} of cases, beating both Co-Attack \textbf{39.47\%} and SGA \textbf{17.84\%}, proving ViPro's stealthiness.

\subsection{Discussion}
\label{sec:discussion}
In this section, we will elaborate on the mechanism behind the ViPro attack, including a qualitative analysis on the determinants for \textbf{upper bound} and the \textbf{lower bound} of video promotion attacks. We then investigate the potential defensive practices against ViPro attacks. Finally, we discuss the limitations and future work of our paper.

\begin{figure}[t!]
\centering
\includegraphics[width=0.8\linewidth]{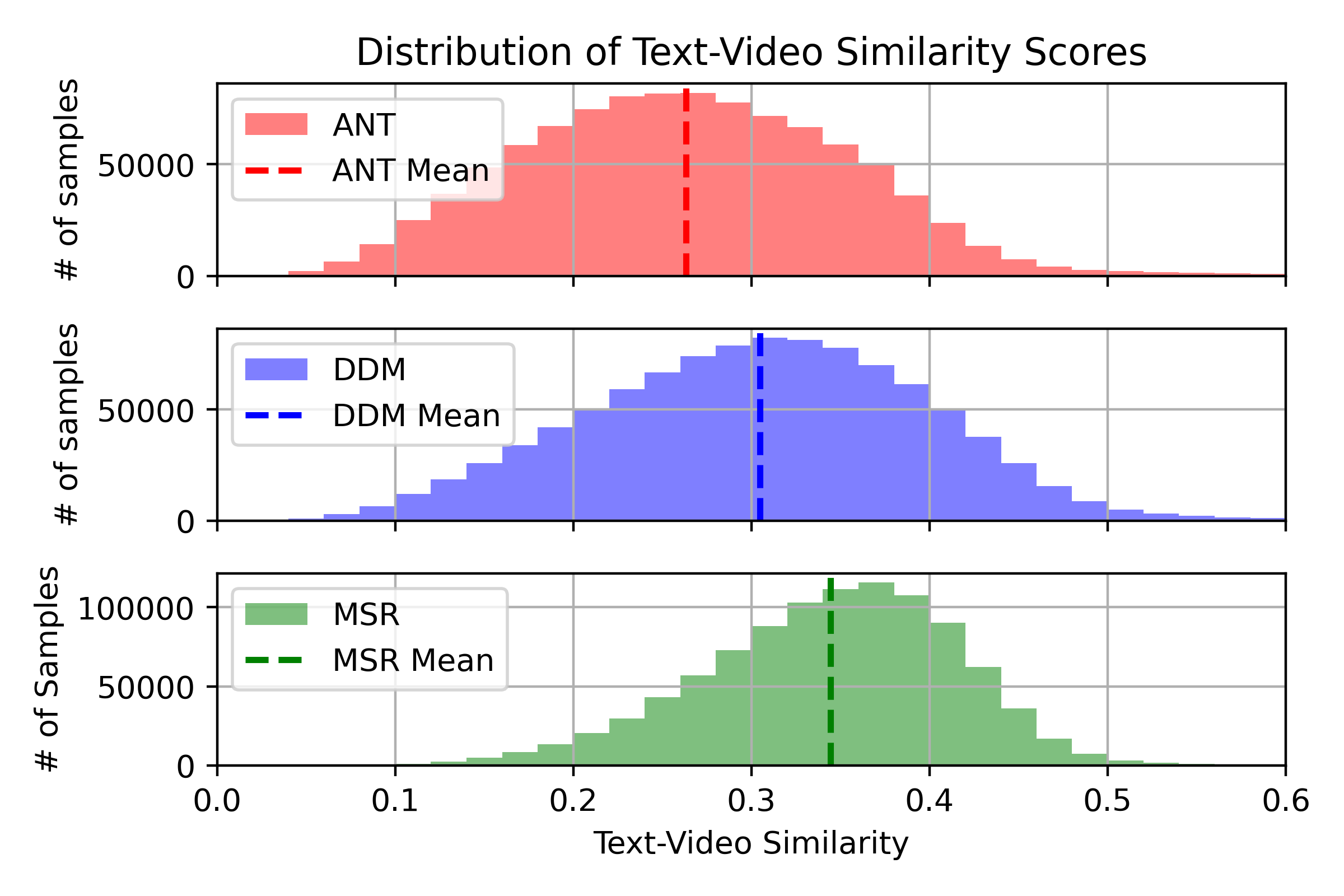}
  \caption{Visualization of white-box top 1 text-to-video similarity scores for all datasets. All results are normalized between 0 and 1 for better visibility.}
\label{fig:dist_d}
\end{figure}

\begin{figure}[h]
\centering
\includegraphics[width=0.8\linewidth]{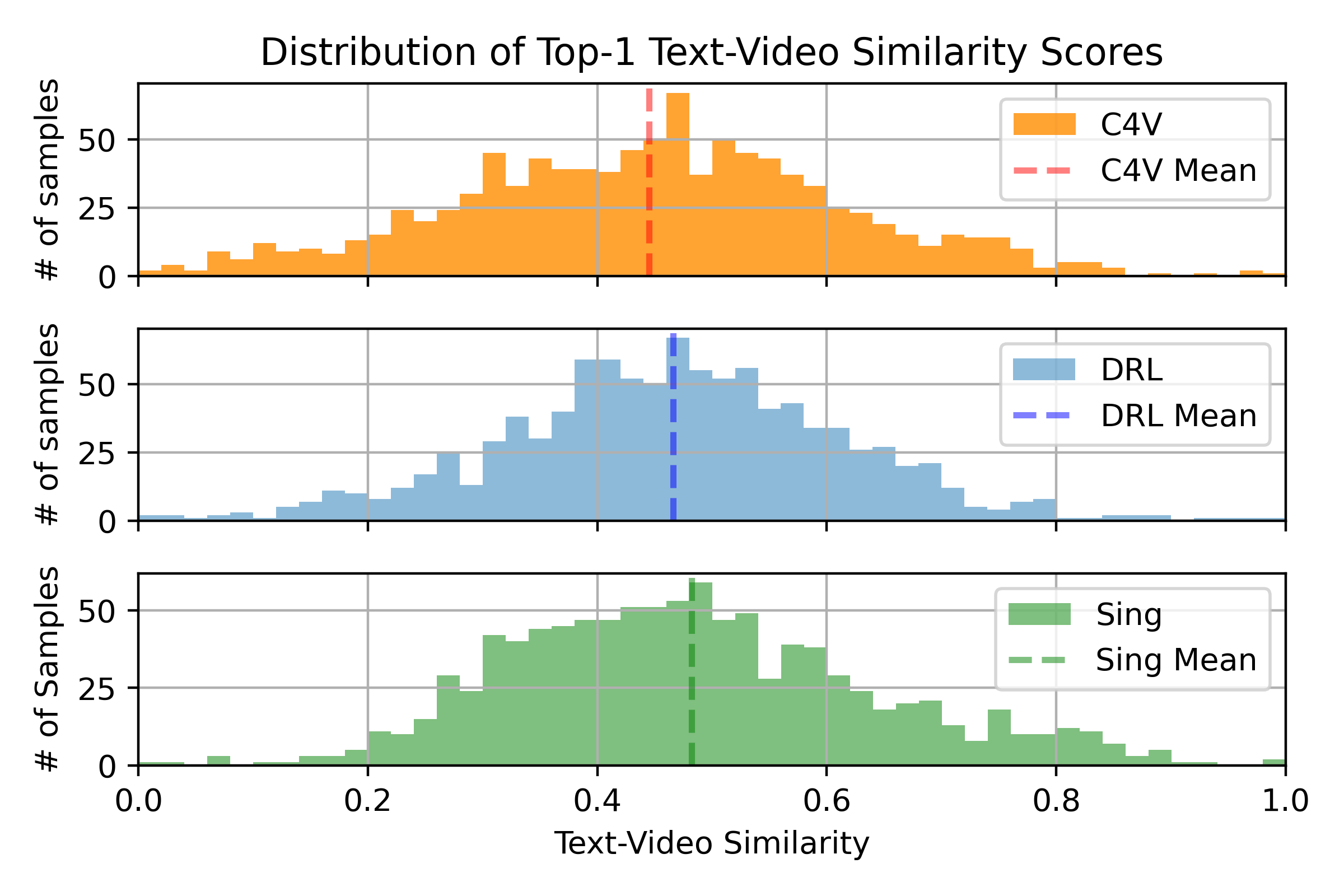}
  \caption{Visualization of white-box top 1 text-to-video similarity scores for all models. Because the embeddings vary significantly across models, we only present normalized top 1 scores for better visibility.}
\label{fig:dist_m}
\end{figure}

\noindent\textbf{Upper Bound.} The upper bound refers to the theoretically achievable maxima on a victim model for all attacks, which is jointly determined by \textit{the distribution of model embeddings} and \textit{the distribution of the dataset}. The results w/o $\epsilon$ across datasets and models in Table.\ref{tabl_vipro_wb_d} and Table.\ref{tabl_vipro_wb_m} can also support the previous conclusion. For example, ActivityNet has the highest R@1 and R@5 than MSR-VTT and DiDeMo, while the results on Sing show the lowest R@1/5 over DRL and C4V. Following our previous theory on the retrieval region, ActivityNet is supposed to have the largest retrieval region for all datasets, while Sing has the tightest (smallest) region for all models. This translates to the smallest average text-video similarities for ActivityNet and the largest ones for Sing. To validate this, we plot histograms of text-video similarities for all datasets and models. As shown in Figure.\ref{fig:dist_d} and Figure.\ref{fig:dist_m}, the distribution and mean of the similarities for all datasets and models align with our previous speculation. The results also echo our previous speculation on model `sensitivity': With the largest average top-1 similarity, Sing \textcolor{black}{exhibits the highest compactness and becomes} the most sensitive model to variation in the embedding, making SGA the least effective attack due to augmentation, and vice versa for C4V and DRL. This verifies our theory on the two determinants: the model distribution of model embeddings and the dataset.

\begin{table}[t!]
     \caption{\label{imper_eval} \textcolor{black}{Image quality evaluation using SSIM, PSNR, and LPIPS for all attacks. Results are given in average for all models.} }
    \centering
   \resizebox{0.8\linewidth}{!}{
   \begin{tabular}{c|c|c|c}
    \toprule
    \textcolor{black}{\textbf{Method}}  & \textcolor{black}{\textbf{SSIM}} $\uparrow$   &  \textcolor{black}{\textbf{PSNR}} $\uparrow$  &\textcolor{black}{\textbf{LPIPS}} $\downarrow$   \\ 
     \cmidrule{1-4}     
  \textcolor{black}{Co-Attack}  \cite{Coattack}  &  \textcolor{black}{0.945}        &    \textcolor{black}{33.81}   &   \textcolor{black}{0.043} \\
  \textcolor{black}{SGA} \cite{SGA}     &   \textcolor{black}{0.940}        &    \textcolor{black}{33.33}   &  \textcolor{black}{0.048} \\
  \textcolor{black}{ViPro}             &  \textcolor{black}{ 0.944}        &    \textcolor{black}{33.60}   &   \textcolor{black}{0.045} \\
    \bottomrule
    \end{tabular}}
\end{table}

\noindent\textbf{Lower Bound.} The lower bound refers to the generalization ability of attacks to maintain their effectiveness across settings. We hypothesize the algorithms of attacks and the architecture of victim models as the two determining factors. For algorithms, we show in the main results and ablation studies that ViPro consistently outperforms its counterparts using KL divergence (Co-Attack) and naive negative loss (SGA). Besides, the incorporation of MoRe also implies that the explicit design in refining temporal and semantical interaction further improves the ability to generalize across models. As for model architecture, we find that models with a structural cross-modal interaction exhibit stronger resistance to non white-box attacks. For example, as shown in Table.\ref{tabl_vipro_gb_m}, DRL experiences the least performance drop compared to white-box results, given that it is the only model using algorithmic cross-modal interaction. C4V, in contrast, shows the strongest resistance because of its video aggregator within the cross-modal module. Furthermore, according to our ablation study in the Supplementary, the inclusion of access to the video aggregator would significantly boost all attacks compared to grey-box settings. These findings shed light on possible ways in building robust T2VR models, especially when using open-sourced pre-trained encoders.

\noindent\textbf{\textcolor{black}{Attacks on Commercial Platforms.}}
\textcolor{black}{
Successfully attacking commercial two-stage T2VR platforms, i.e., YouTube/TikTok, could still be challenging. This is because real-world platforms rely more than simply visual information, i.e., user patterns, metadata of the video (clicks, watches) and transcriptions of the soundtrack This diverse modality reliance largely alleviates the impact of visual perturbation introduced by ViPro. Hence, ViPro is more likely to influence the retrieval stage in cold-start where visual features play a dominant role, on the premise of optimizing ViPro on backbones with similar scales and architectures. Thus, while we did not explicitly validate ViPro on real-world platforms, it keeps threatening the secure application of T2VR and calling for defensive counterplays. }

\noindent\textbf{Potential Defense against ViPro.}
As presented in Table.\ref{tabl_wb_def} and Table.\ref{tabl_gb_def}, ViPro remains effective under popular image and video defences. To better protect T2VR models from such attacks, popular methods such as adversarial fine-tuning and purification could also be an efficient protection. However, the former may cause large performance degradation, while the latter requires computational optimization for faster video purification. Thus, we suggest using more information-rich modalities, such as audio, for multimodal fusion. From our observation for the attack lower bound, the inclusion of encoders for more modalities would increase the robustness of models and weaken the dependence of models on visual information.

\noindent\textbf{Limitations \& Future Work.}
Due to the computational overhead, we cannot test our attacks on larger VLMs but use deployable open-sourced models for evaluations. Thus, we cannot launch effective attacks on larger commercial T2VR platforms such as YouTube. Besides, the transferability can be further boosted using more accurate temporal clipping, i.e., model-aided clustering, and/or model ensemble. Lastly, temporal constraints can also be applied for better stealthiness of manipulated videos to avoid visual clues and artifacts. These directions are valuable for future exploration.
\section{Conclusion}
We explore the overlooked vulnerability against text-to-video retrieval (T2VR) and propose a new attacking paradigm to promote video ranks adversarially. Accordingly, we pioneer the Video Promotion attack (ViPro) as the first attack targeting such vulnerability. We further propose Modality Refinement (MoRe) to capture the intricate modality interaction to enhance black-box transferability. Comprehensive experiments include \textbf{2} existing baselines, \textbf{3} leading T2VR models, \textbf{3} prevailing datasets with over 10k videos, evaluated under \textbf{3} scenarios. All experiments are conducted in a multi-target setting to reflect realistic scenarios where attackers seek to promote the video regarding multiple queries simultaneously. We also provide evaluations for defences and imperceptibility. In sum, ViPro consistently shows the best performance on all settings over existing baselines, implying the superiority of our method. Our work provides a qualitative analysis on the upper and lower bounds of ViPro, highlights an overlooked vulnerability, and offers insights into potential counterplays. 

\section*{Acknowledgement}
This work was supported by National Key Research and Development Program of China (2023YFB3107401), the National Natural Science Foundation of China (T2341003,
62521002, U2441240, U24B20185, 62376210, 62132011, 62406240).

\bibliographystyle{IEEEtranN}
\bibliography{./reference.bib}

\begin{IEEEbiography}
[{\includegraphics[width=1in,height=1.2in,clip,keepaspectratio]{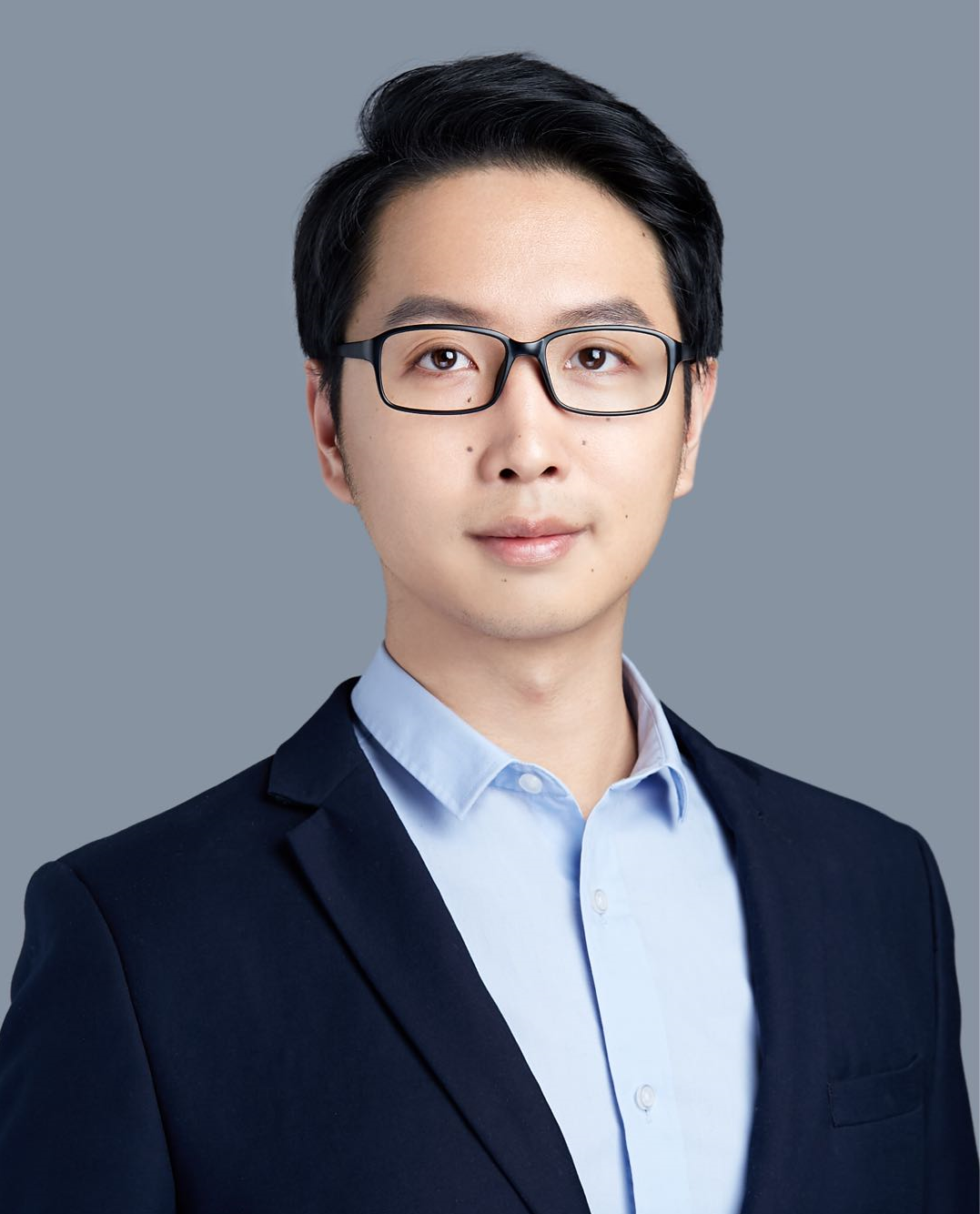}}]{Qiwei Tian} received his B.E. in automation from Tianjin University in 2014, his M.Sc. degree in electrical engineering from the University of Melbourne in 2017. He is now a Ph.D. student in Cyber Science and Engineering at Xi'an Jiaotong University. His major scopes of research include adversarial attacks and defense, multimodal retrieval, and Vision-Language models.
\end{IEEEbiography}

\begin{IEEEbiography}
[{\includegraphics[width=1in,height=1.25in,clip,keepaspectratio]{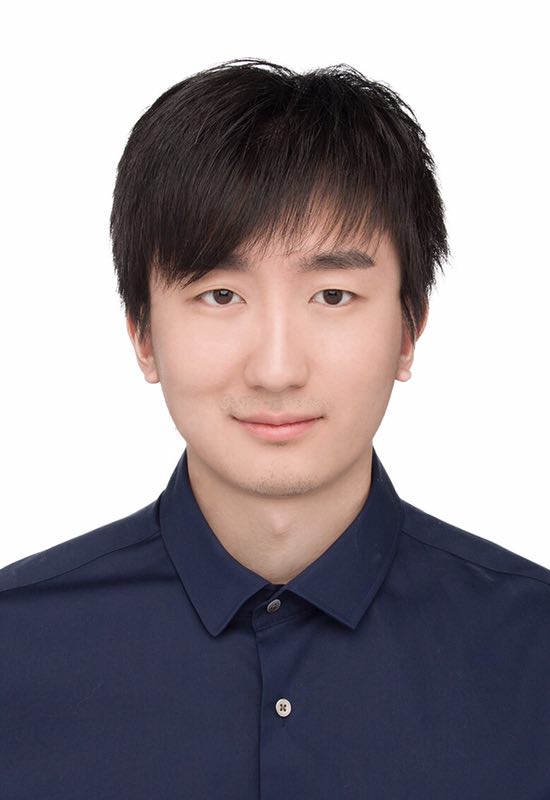}}]
{Chenhao Lin} (Member, IEEE) received the B.Eng. degree in automation from Xi'an Jiaotong University in 2011, the M.Sc. degree in electrical engineering from Columbia University in 2013, and the Ph.D. degree from The Hong Kong Polytechnic University in 2018. He is currently a Professor at Xi'an Jiaotong University of China. His research interests are in artificial intelligence security, identity authentication, biometrics, adversarial attack and robustness, and pattern recognition.
\end{IEEEbiography}

\begin{IEEEbiography}
[{\includegraphics[width=1in,height=1.25in,clip,keepaspectratio]{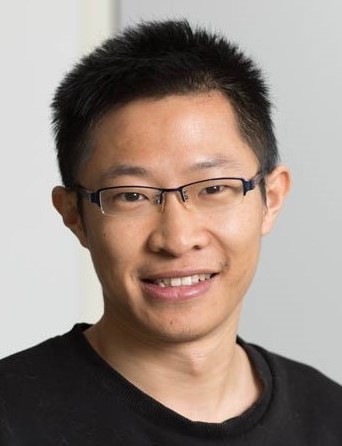}}]
{Zhengyu Zhao} (Member, IEEE) received the Ph.D. degree from Radboud University, The Netherlands. He is currently a Professor at Xi'an Jiaotong University, China. His research focuses on adversarial machine learning, covering its foundations (in adversarial examples and data poisons) and applications (in multi-modal generative models and autonomous driving).
\end{IEEEbiography}

\begin{IEEEbiography}
[{\includegraphics[width=1in,height=1.2in,clip,keepaspectratio]{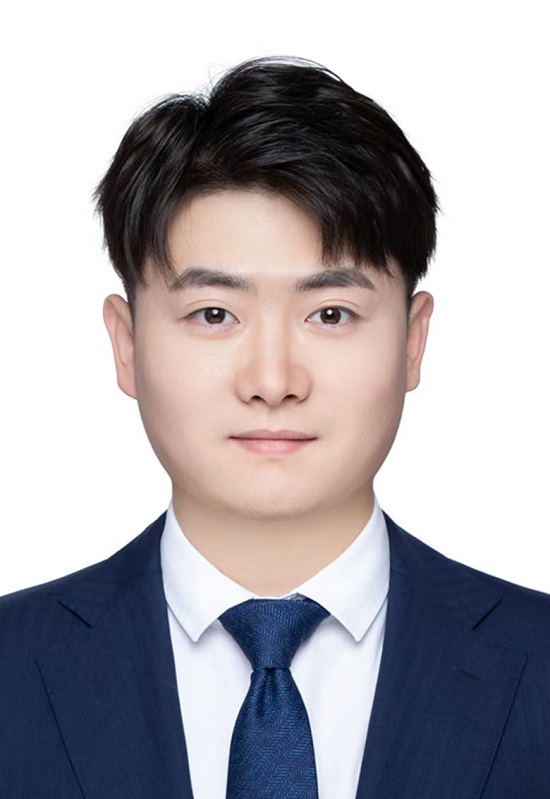}}]
{Qian Li} (Member, IEEE) received the Ph.D. degree in computer science and technology from Xi'an Jiaotong University, China, in 2021, where he is currently an assistant professor with the School of Cyber Science and Engineering in Xi'an Jiaotong University, China. His research interests include artificial intelligence security, trustworthy machine learning and deep adversarial learning, with an emphasis on robustness, privacy, generalization, and their interconnections.
\end{IEEEbiography}

\begin{IEEEbiography}
[{\includegraphics[width=1in,height=1.25in,clip,keepaspectratio]{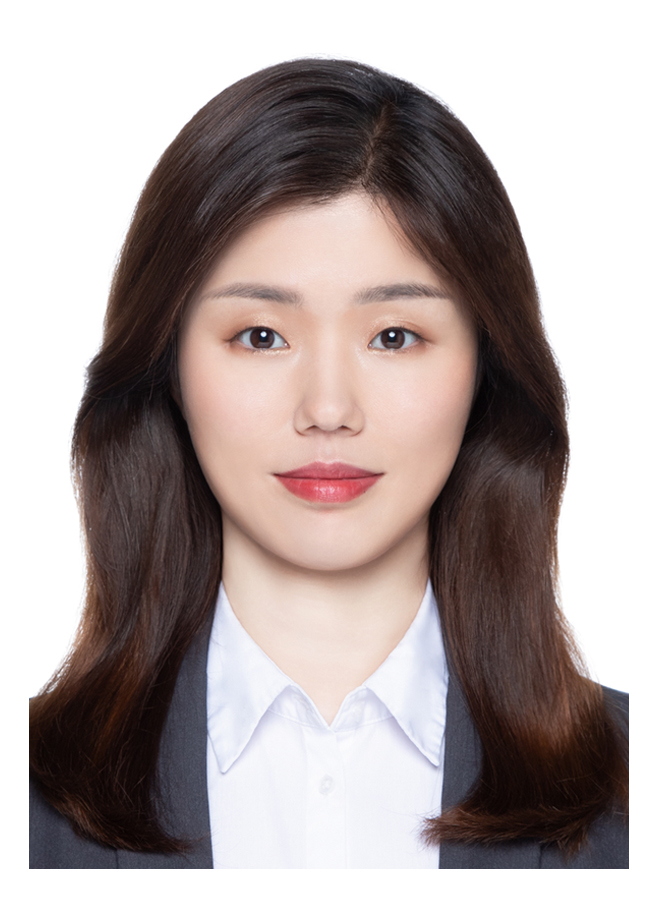}}]
{Shuai Liu}  (Member, IEEE) received the B.S. degree in Control Science and Engineering and Ph.D. degree in Electronic Science and Technology from Xidian University, Xi'an, China, in 2009 and 2017, respectively. She is currently an associate professor at the School of Software Engineering, Xi'an Jiaotong University, Xi'an, China. Her major research interests include computer vision, deep learning, large language models, and artificial intelligence security
\end{IEEEbiography}
\vspace{11pt}

\begin{IEEEbiography}
[{\includegraphics[width=1in,height=1.25in,clip,keepaspectratio]{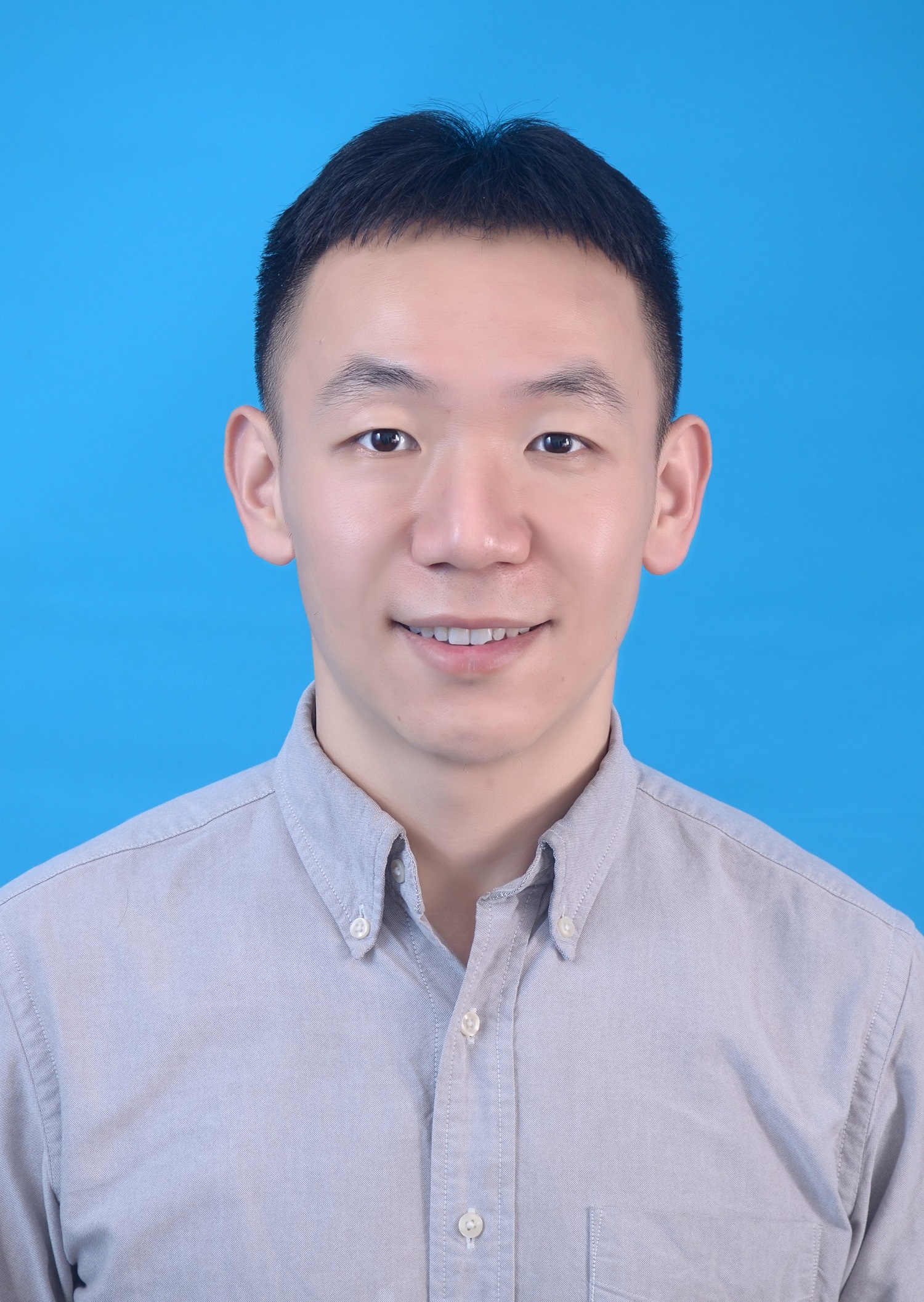}}]
{Chao Shen} (Fellow, IEEE)
received the B.S. degree in Automation from Xi'an Jiaotong University, China in 2007; and the Ph.D. degree in Control Theory and Control Engineering from Xi'an Jiaotong University, China, in 2014. He is currently a Distinguished Professor in the Faculty of Electronic and Information Engineering, Xi'an Jiaotong University of China. His current research interests include AI Security, insider/intrusion detection, behavioral biometrics, and measurement and experimental methodology.
\end{IEEEbiography}

\clearpage
\appendix

\setcounter{page}{1}
\appendix

\begin{figure}[t]
\centering
\includegraphics[width=\linewidth]{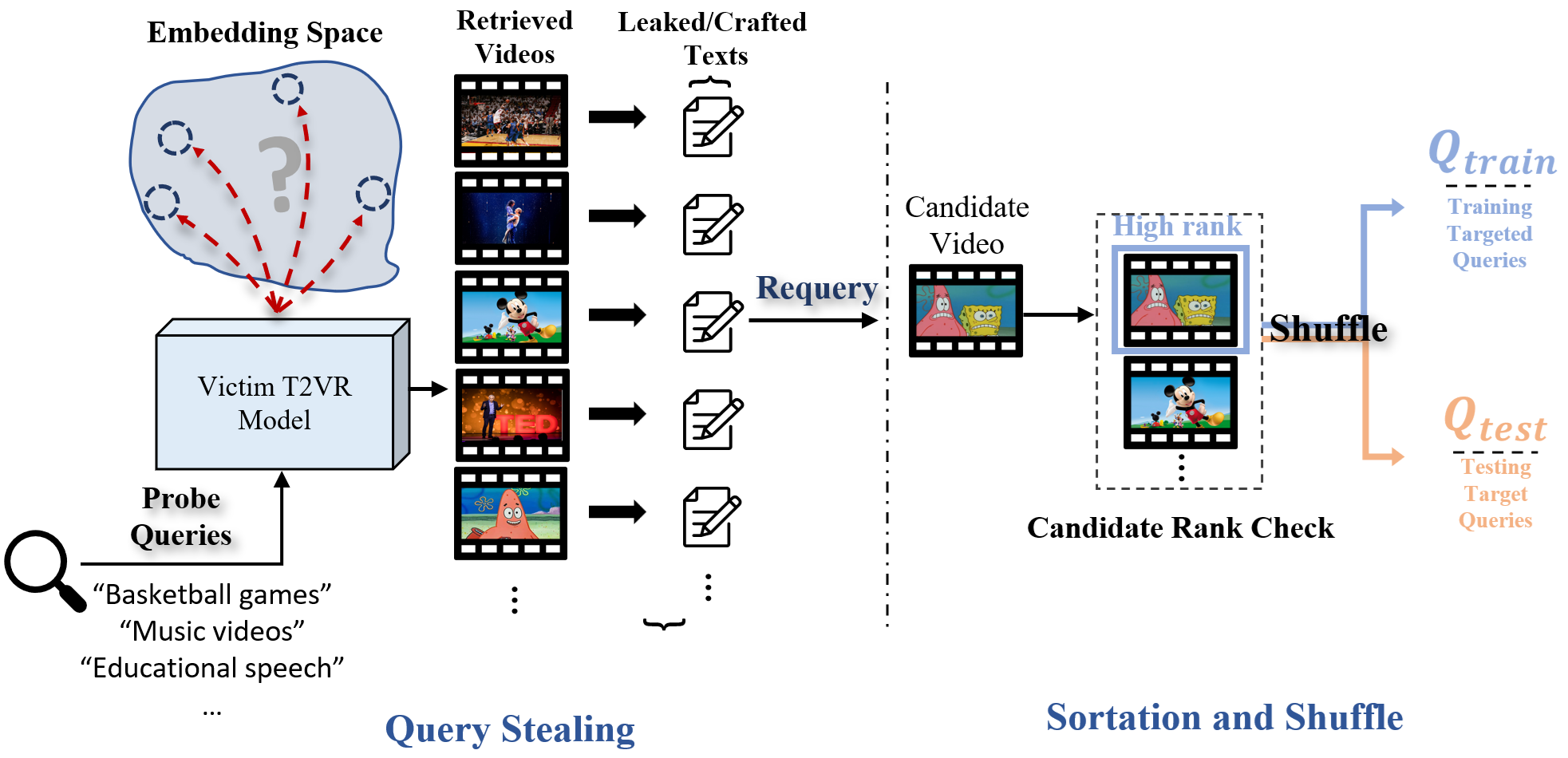}
  \caption{\label{fig:de}
An illustration of data stealing: \textit{i}. Attackers can obtain a diverse subset of training videos and corresponding texts by querying the victim model. \textit{ii}. Attackers can perform a candidate-wise sorting for \textbf{each} candidate video by querying the candidate video with the texts from step \textit{i}, checking the rank of the candidate video, and categorizing the texts into \textbf{Relevant} and \textbf{Irrelevant}. }
\end{figure}

\noindent\textbf{Data Construction.}
As mentioned in the main paper, we find targeted queries of the ViPro attack by conducting a candidate-wise sorting, as shown in Fig.\ref{fig:de}. Specifically, these texts are re-ranked by using them to re-query the model and check the rank of the candidate under these texts. If the candidates are within the top-20 results, the corresponding captions are categorized as targeted queries. The re-ranked 20 texts are then shuffled and evenly divided into train sets and test sets. \textit{Re-rank are conducted using the Singularity-17M model.}

\begin{figure*}[t]
\centering
\includegraphics[width=\linewidth]{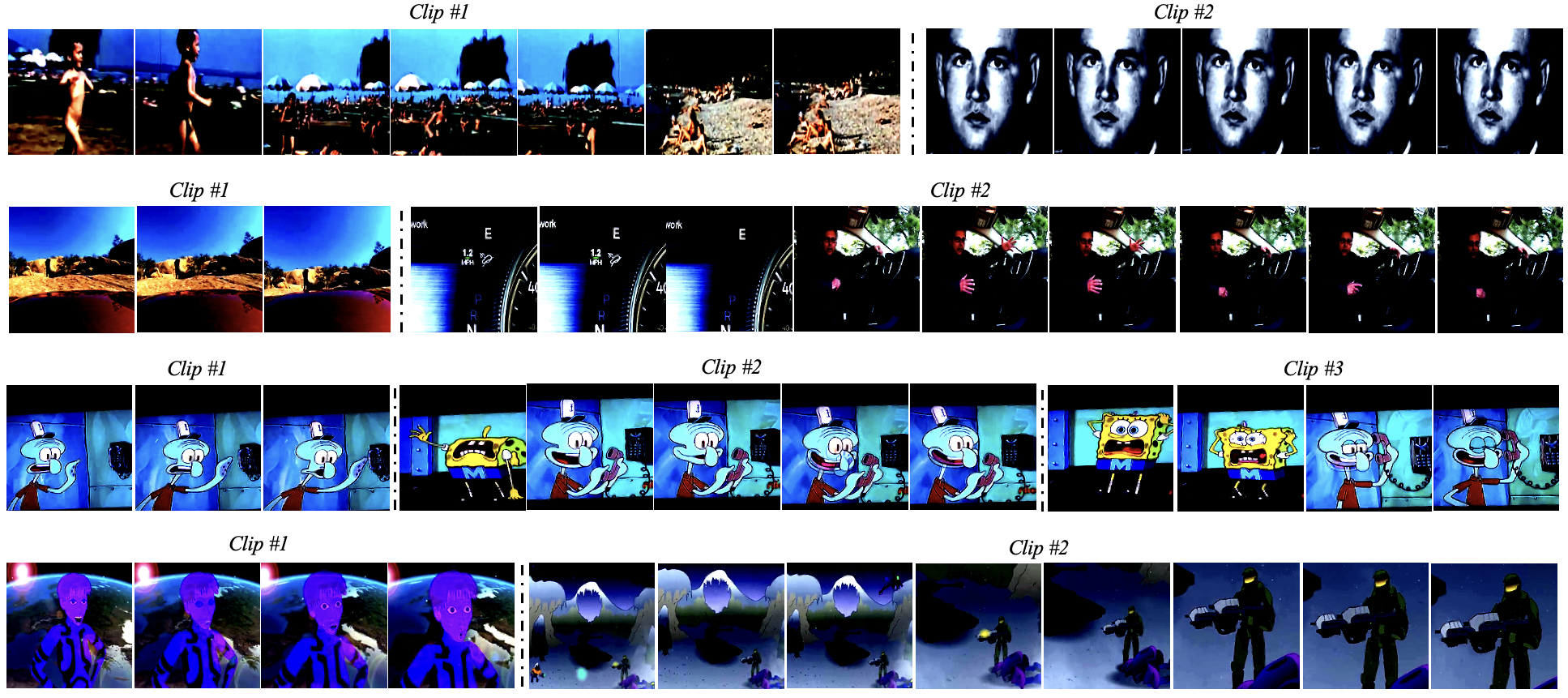}
  \caption{\label{fig:clip_vl}
An illustration of the clipped video frames through temporal clipping. All temporally related frames are grouped into clips without creating too fragmented videos, i.e., only 1-2 frames per clip.}
\end{figure*}

\noindent\textbf{Pseudo-code for Temporal Clipping (TC).}
We provide pseudo-code for Temporal Clipping in Alg.\ref{alg_vc}. TC first calculates the frame-to-frame correlation using cosine similarity to get $\mathbf{W}_{\mathbf{X}}$, and then calculates the temporal difference between frames $\Delta_{\mathbf{W}_\mathbf{X}}$. Frames that exceed $\gamma$ will be marked as temporally outlying frames and clipped if the clip length is longer than $T/4$, which ensures that the video is not too fragmented to preserve temporal information. We also provide an illustration on the result of TC, as shown in Fig.\ref{fig:clip_vl}.

\noindent\textbf{Categorical Performance.}
As discussed in the main paper, we use the categorical information provided in \cite{ref:MSRVTT} and choose the top 10 categories with the best R-Precisions. We provide the R-Precisions for all models on all categories as follows \textcolor{black}{in Table.\ref{tabl_vipro_gb_m} and Table.\ref{tabl_vipro_cat}}. Interestingly, we find that all categories with low R-Precision are very non-specific words, such as ``documentary'', ``how-to'' and ``people'', with which VLMs might find it difficult to relate visual contents. On the other hands, categories with higher precisions are generally ones with concrete implications, such as ``animation'', ``sports'', and ``gaming''. For the distribution of data in each category, please refer to the original paper.
Based on the R-Precision, we chose the 10 categories with the highest average performance: Sports, Movie, Animation, Music, Animal, Vehicles, Gaming, News, Education, Cooking.

\noindent\textbf{\textcolor{black}{Impact of Video Aggregator.}}
\textcolor{black}{Finally, we evaluate the impact of the access to the video aggregator on the C4V model. As shown in Table.\ref{tabl_c4v_agg}, all attacks gain significant performance boost after gaining the access to the video aggregator, validating the significance of such access for generating efficient grey-box attacks. Furthermore, our ViPro retains its lead among all other methods by over 50\% R@1 boosts.}

\begin{algorithm}[t!]
\caption{Pseudo-code for Temporal Clipping.}
\algrenewcommand\algorithmicrequire{\textbf{\textcolor{black}{Input:}}}
\algrenewcommand\algorithmicensure{\textbf{\textcolor{black}{Output:}}}
\label{alg_vc}
\begin{algorithmic}[1]
    
\Require Clean video $\mathbf{X}=[x_1,...,x_T]$, video features $\mathbf{F}_{\mathbf{X}}$ threshold for outlying frames $\gamma$
\Ensure Clipped Videos $\mathbb{C}$
\Function{TempClip}{$\mathbf{W}_\mathbf{X},\mathbf{X}$}     
    \State $\mathbf{W}_{\mathbf{X}}\gets$ \Call{CosSim}{$\mathbf{F}_{\mathbf{X}},\mathbf{F}_{\mathbf{X}}$}  \Comment{$T\times T$ temporal similarity}
    \State $row\gets0$, $\Delta_{\mathbf{W}_\mathbf{X}} \gets$\Call{GetDiff}{$\mathbf{W}_\mathbf{X}$} 
    \While{$row<T$}
        \State $col\gets row$
        \While{$col<T$}
            \State $\Delta\gets \Delta_{\mathbf{W}_\mathbf{X}}[row,col]$
            \If {$\Delta\geq\gamma$} and $col\geq T/4$
                \State Clip the similarity $\mathbf{W}_\mathbf{X}$ and the video $\mathbf{X}$
                \State Break
            \Else
                \State $col\gets col+1$
            \EndIf
        \EndWhile
        \State $row\gets row+1$
    \EndWhile
    \State \textbf{return} Clipped Videos $\mathbb{C}$
\EndFunction
\end{algorithmic}
\end{algorithm}

\begin{table*}[t!]
    \footnotesize
    \tiny
     \caption{\label{tabl_vipro_gb_m} R-Precision for \textbf{first 10} categories on all models. The datset is MSR-VTT-1k. Chosen categories are highlighted in \textbf{bold}.}
    \centering
   \resizebox{\linewidth}{!}{
   \begin{tabular}{c|c|c|c|c|c|c|c|c|c|c}
    \toprule
   \multirow{2}*{\textbf{Models}} &  \multicolumn{10}{c}{\textbf{R-Precision (\%)}} \\
   \cmidrule{2-11}
                      &  \textbf{Music} & People & \textbf{Gaming} & \textbf{Sports} & News & \textbf{Education} & TV shows & \textbf{Moive} & \textbf{Animation} & \textbf{Vehicles} \\
     \cmidrule{1-11}
    {Sing} &  35.14 &  0.00     &   24.53      &   42.11         & 13.79  & 15.38 &    20.00 & 31.15  & 26.32   & 25.00         \\
     \cmidrule{1-11}
                {DRL}  &  44.59  &  0.00     &  37.74      &      64.21     &29.31    &34.62   &16.67    & 49.18   &36.84   &   51.32     \\
    \cmidrule{1-11}
                {C4V}  &   35.14  &   0.00   &   35.85    &       46.32      & 53.45   & 42.31   &26.67  & 44.26    &57.89   &   23.68     \\
              
    \bottomrule
    \end{tabular}}

\end{table*}

\begin{table*}[t!]
    \footnotesize
    \tiny
     \caption{\label{tabl_vipro_cat} R-Precision for \textbf{last 10} categories on all models. The dataset is MSR-VTT-1k. Chosen categories are highlighted in \textbf{bold}.}
    \centering
   \resizebox{\linewidth}{!}{
   \begin{tabular}{c|c|c|c|c|c|c|c|c|c|c}
    \toprule
   \multirow{2}*{\textbf{Models}} &  \multicolumn{10}{c}{\textbf{R-Precision (\%)}} \\
   \cmidrule{2-11}
                      & How-to & Travel & Science &\textbf{Animal} & Kids & Documentary & \textbf{Cooking} & food & Beauty & Advertisement \\
     \cmidrule{1-11}
    {Sing} &  13.04   & 3.45     &   11.63      &  12.07         & 10.64  &0.00 &    12.50 & 9.09  & 11.32   & 12.50         \\

     \cmidrule{1-11}
{DRL}  &  13.04   &  24.14     &   27.91      &   55.17        & 25.53  & 0.00 &    37.50 & 45.45  & 11.32   & 16.67         \\

     \cmidrule{1-11}   
{C4V}  &  13.04  &  31.03   &     34.88    &     37.93      & 8.51  &  0.00 &    15.62 & 33.33  & 32.08  &  20.83     \\
              
    \bottomrule
    \end{tabular}}

\end{table*}

\begin{table*}[t!]
    \footnotesize
    \tiny
     \caption{\label{tabl_c4v_agg} An ablation study on the influence of the access to the video aggregating transformer within the C4V model. \textbf{All attacks have gained a performance boost with access to the aggregator.} Our ViPro maintains its lead in either scenario.}
    \centering
   \resizebox{\linewidth}{!}{
   \begin{tabular}{c|c|c|c|c}
    \toprule
    {\textbf{Method}}  &  Access to Video Aggregator &  \textbf{R@1}(\%) $\uparrow$  & \textbf{R@5}(\%) $\uparrow$ & \textbf{Average} (\%) $\uparrow$  \\ 
     \cmidrule{1-5}     
 \multirow{2}*{Co-Attack} &  \ding{55}        &    \textcolor{black}{0.93 (-2.83)}      &   \textcolor{black}{5.07 (-8.35) }     &     \textcolor{black}{3.00 (-5.59)}              \\
    \cmidrule{2-5}  
                            & \checkmark      &    26.48 (+22.72) &  53.11 (+39.69) & 43.05 (+32.21) \\
     \cmidrule{1-5}     
 \multirow{2}*{SGA} &  \ding{55}       &      \textcolor{black}{1.28 (-2.48)}      &   \textcolor{black}{7.69 (-5.73)}      &       \textcolor{black}{4.49 (-4.11)}               \\
    \cmidrule{2-5}  
                            & \checkmark      &   { 5.20 (+1.44)} & 18.98 (+5.56) & 12.09 (+3.50) \\
     \cmidrule{1-5}     
 \multirow{2}*{ViPro} &  \ding{55}       &   \textcolor{black}{1.99 (-1.77)}      &  \textcolor{black}{10.53 (-2.90)}      &        \textcolor{black}{6.26 (-2.34)    }        \\
    \cmidrule{2-5}  
                            & \checkmark      &    \textbf{77.10 (+73.34)}      &  \textbf{90.65 (+77.23)}      &        \textbf{83.88 (+75.29)    }        \\
    \bottomrule
    \end{tabular}}

\end{table*}

\end{document}